\documentclass[journal]{IEEEtran}
\usepackage{xcolor,soul,framed,caption} 

\colorlet{shadecolor}{yellow}
\usepackage[pdftex]{graphicx}
\graphicspath{{../pdf/}{../jpeg/}}
\DeclareGraphicsExtensions{.pdf,.jpeg,.png}

\usepackage[cmex10]{amsmath}

\usepackage{array}
\usepackage{mdwmath}
\usepackage{mdwtab}
\usepackage{eqparbox}
\usepackage{url}
\usepackage{supertabular}                        \usepackage[colorlinks=true,
    linkcolor=blue,
    urlcolor=blue,
    citecolor=blue]{hyperref}    
\usepackage{amsmath,amssymb}
\usepackage{caption}
\usepackage{subcaption}
\usepackage{float}
\usepackage{booktabs}
\usepackage{multirow}
\usepackage{graphicx}
\usepackage{color}
\usepackage{tabularray}
\usepackage{hhline}
\usepackage{wasysym}
\usepackage[ruled, lined, longend, linesnumbered]{algorithm2e}
\usepackage{algorithmic}
\usepackage{tikz}
\usetikzlibrary{positioning,fit,shapes.misc,arrows.meta,calc}

\usepackage{setspace}

\usepackage{titlesec}
\titlespacing*{\section}{0pt}{0.5\baselineskip}{0.3\baselineskip}
\titlespacing*{\subsection}{0pt}{0.4\baselineskip}{0.2\baselineskip}
\titlespacing*{\subsubsection}{0pt}{0.3\baselineskip}{0.1\baselineskip}


\begin{document}
\bstctlcite{IEEEexample:BSTcontrol}
    \title{DeFRiS: Silo-Cooperative IoT Applications Scheduling via Decentralized Federated Reinforcement Learning}
  \author{Zhiyu Wang, Mohammad Goudarzi, Mingming Gong, and Rajkumar Buyya
  \thanks{Zhiyu Wang and Rajkumar Buyya are with The Quantum Cloud Computing and Distributed Systems (qCLOUDS) Laboratory, School of Computing and Information Systems, The University of Melbourne, Australia (e-mail: zhiywang1@student.unimelb.edu.au, rbuyya@unimelb.edu.au).}
  \thanks{Mingming Gong is with School of Mathematics and Statistics, The University of Melbourne, Australia (email: mingming.gong@unimelb.edu.au).}
  \thanks{Mohammad Goudarzi is with The Faculty of Information Technology, Monash University, Australia (email: mohammad.goudarzi@monash.edu).}
}  

\maketitle

\begin{abstract}
Next-generation Internet of Things (IoT) applications increasingly span across autonomous administrative entities, necessitating silo-cooperative scheduling to leverage diverse computational resources while preserving data privacy. However, realizing efficient cooperation faces significant challenges arising from infrastructure heterogeneity, Non-Independent and Identically Distributed (Non-IID) workload shifts, and the inherent risks of adversarial environments. Existing approaches, relying predominantly on centralized coordination or independent learning, fail to address the incompatibility of state-action spaces across heterogeneous silos and lack robustness against malicious attacks. To address these challenges, this paper proposes DeFRiS, a Decentralized Federated Reinforcement Learning framework for robust and scalable Silo-cooperative IoT application scheduling. DeFRiS integrates three synergistic innovations: (i) an action-space-agnostic policy utilizing candidate resource scoring to enable seamless knowledge transfer across heterogeneous silos; (ii) a silo-optimized local learning mechanism combining Generalized Advantage Estimation (GAE) with clipped policy updates to resolve sparse delayed reward challenges; and (iii) a Dual-Track Non-IID robust decentralized aggregation protocol leveraging gradient fingerprints for similarity-aware knowledge transfer and anomaly detection, and gradient tracking for optimization momentum. Extensive experiments on a distributed testbed with 20 heterogeneous silos and realistic IoT workloads demonstrate that DeFRiS significantly outperforms state-of-the-art baselines. Specifically, DeFRiS reduces average response time by 6.4\% and energy consumption by 7.2\% compared to the best-performing baseline, while lowering tail latency risk (CVaR$_{0.95}$) by 10.4\% and achieving near-zero deadline violations. Furthermore, extensive experiments reveal that DeFRiS achieves over 3 times better performance retention as the system scales and over 8 times better stability in adversarial environments compared to the best-performing baseline.
\end{abstract}

\begin{IEEEkeywords}
Internet of Things, Edge/Cloud Computing, Deep Reinforcement Learning, Federated Learning
\end{IEEEkeywords}

\IEEEpeerreviewmaketitle

\section{Introduction}
\IEEEPARstart{T}he Internet of Things (IoT) has become ubiquitous in smart manufacturing, healthcare, and urban infrastructure \cite{chang2024digital}. Traditional deployments strictly adhere to single-authority boundaries, where a solitary entity retains exclusive control over resource provisioning and scheduling \cite{taleb2025survey}. However, next-generation applications are breaking these boundaries, demanding cross-organizational cooperation \cite{wu2024federated}. In smart cities, distinct departments manage traffic, energy, and emergency services utilizing separate infrastructures that must nevertheless coordinate during critical events. Similarly, industrial supply chains link manufacturers, logistics providers, and retailers, necessitating computational synergy among independent systems without exposing proprietary data. These scenarios share a fundamental requirement: enabling resource cooperation across autonomous administrative boundaries while maintaining strict data privacy. We term each such independent administrative unit a computing silo \cite{zhang2024high}.

Realizing multi-silo cooperation, however, presents distinct technical challenges. First, significant infrastructure heterogeneity exists across silos tasked with scheduling applications modeled as Directed Acyclic Graphs (DAGs) \cite{li2025dynamic}. Mapping these complex task dependencies onto diverse hardware configurations creates severe compatibility issues. A silo equipped with sparse edge resources and abundant cloud capacity presents a fundamentally different state-action space compared to an edge-dense counterpart. This dimensional and semantic mismatch makes transferring scheduling policies between silos non-trivial. Second, workload variations exacerbate this complexity \cite{10887320}. The divergence between real-time stream processing and batch analytics creates a Non-Independent and Identically Distributed (Non-IID) shift in underlying data distributions \cite{fan2024taking, uddin2025systematic}. Indiscriminate aggregation of knowledge across such diverse environments can induce negative transfer, where incompatible patterns degrade rather than enhance performance. Finally, the scheduling environment involves inherent uncertainties and adversarial risks \cite{10623738}. The optimization process is plagued by sparse, delayed rewards, where feedback spans hundreds of decision steps \cite{yunis2024subwords}, and is further threatened by potential hardware failures or malicious model poisoning \cite{li2023end}.

Deep Reinforcement Learning (DRL) provides a powerful solution for addressing these dynamic scheduling complexities \cite{10417755}, yet its application in multi-silo environments faces a fundamental dilemma between autonomy and cooperation. Existing approaches generally adopt either independent learning or cooperative learning paradigms. Independent learning preserves data sovereignty by restricting training to local resources, but this isolation precludes the sharing of optimization experience, resulting in severe sample inefficiency and suboptimal convergence \cite{qiu2020distributed}. Conversely, cooperative learning seeks to synthesize collective intelligence but faces distinct structural and algorithmic hurdles \cite{jin2024collaborative}. Conventional cooperative frameworks typically rely on centralized coordination mechanisms that introduce communication bottlenecks and single points of failure, effectively compromising scalability and robustness \cite{gecer2024federated}. More critically, these approaches predominantly operate under idealized assumptions of model homogeneity and truthful participation \cite{ye2023heterogeneous}. Consequently, they lack the architectural flexibility to support heterogeneous state-action spaces, the mathematical robustness to mitigate Non-IID distribution shifts, and the security mechanisms to withstand adversarial environments (e.g., malicious nodes, hardware failures, network disruptions) \cite{gecer2024federated, ye2023heterogeneous, wu2024topology}.

To bridge this gap, we propose a Decentralized Federated Reinforcement Learning framework for Silo-Cooperative IoT Application Scheduling (DeFRiS). DeFRiS fundamentally dismantles the centralized bottleneck by adopting a fully decentralized peer-to-peer architecture, where silos exchange knowledge via dynamically selected neighbors without a central coordinator. The framework integrates three synergistic innovations to resolve the aforementioned technical hurdles. First, to overcome infrastructure heterogeneity, we propose a novel action-space-agnostic policy. By employing candidate resource scoring that decouples policy parameters from specific physical dimensions, it enables seamless knowledge transfer across silos with varying configurations. Second, to handle sparse delayed rewards, we design a silo-optimized local learning mechanism. This combines an Actor-Critic architecture with Generalized Advantage Estimation (GAE) to transform sparse terminal signals into dense credits, utilizing clipped policy updates to guarantee training stability. Finally, to ensure robustness against Non-IID shifts and adversarial threats, we introduce a Dual-Track Non-IID robust decentralized aggregation protocol. This employs a dual-track strategy: using gradient fingerprints to detect anomalies and selectively transfer Actor knowledge based on similarity, while leveraging gradient tracking for Critic networks to harmonize global optimization momentum with local value estimation.

The main contributions of this paper are:
\begin{itemize} 
\item We formulate silo-cooperative IoT scheduling as a risk-sensitive joint optimization problem. By incorporating Conditional Value at Risk (CVaR) to manage tail risks alongside energy objectives, we strictly enforce hard constraints while optimizing soft goals, formalizing the system as a distributed Markov Decision Process (MDP).

\item We propose a novel action-space-agnostic policy that decouples decision-making from physical resource dimensions. Utilizing a candidate resource scoring mechanism instead of fixed-dimensional output layers, this approach enables effective policy knowledge sharing across silos with fundamentally heterogeneous resource configurations.

\item We design a silo-optimized local learning mechanism to address the sparse reward challenge in long-horizon scheduling. By integrating GAE for precise credit assignment and clipped policy updates, we ensure training stability and high-quality local parameter generation.

\item We introduce a dual-track Non-IID robust aggregation protocol that enables cooperation and defense. For Actor networks, we employ gradient fingerprint-based similarity measurement to detect anomalies and selectively transfer knowledge; for Critic networks, we utilize gradient tracking to share global optimization momentum while preserving accurate local value estimation.

\item We implement and evaluate DeFRiS on a distributed testbed comprising heterogeneous silos running realistic IoT applications. Extensive experiments demonstrate that DeFRiS significantly outperforms state-of-the-art baselines in terms of response time, energy efficiency, QoS guarantee, scalability, and adversarial robustness. 
\end{itemize}

The rest of the paper is organized as follows. Section~\ref{related_work} reviews related work and identifies research gaps through systematic comparison. Section~\ref{problem} establishes the system model, optimization objectives, and MDP formalization. Section~\ref{framework} details the DeFRiS framework. Section~\ref{evaluation} presents comprehensive experimental evaluation in real heterogeneous environments. Section~\ref{conclusions} concludes and discusses future research directions.

\section{Related Work}
\label{related_work}
This section systematically reviews existing IoT application scheduling approaches, categorizing them into independent and cooperative learning paradigms, and identifies four critical research gaps through a comprehensive qualitative comparison that motivate the design of DeFRiS.

\subsection{Independent Learning-based Approaches}
Chen et al. \cite{10380323} proposed a dependency-aware task offloading approach with Deep Q-Network (DQN) for real-time task offloading in cloud-edge environments. They modeled mobile applications as DAGs and customized DQN to train the decision-making model that considers task parallelism without presetting task priority. The objective is to minimize the response time of mobile applications. Tang et al. \cite{10989563} proposed a redundancy-aware adaptive search offloading approach based on DQN for task offloading in mobile edge computing networks. They designed a fine-grained task recombination scheme to eliminate redundant subtask data and codes, and organized devices into a spatial index MP-tree for efficient search. The objective is to minimize offloading delay and energy consumption. Deng et al. \cite{10848209} proposed a DRL approach integrating Transformer models for DAG task scheduling in the Internet of Vehicles. They employ Transformers to learn vehicle trajectory features, combined with Proximal Policy Optimization (PPO) for decision-making. The objective is to minimize task completion time. Wang et al. \cite{wang2024deep} proposed a DRL-based IoT application scheduling algorithm using PPO for DAG-based IoT applications in fog computing environments. They designed a weighted cost model considering task dependencies. The objective is to optimize load balancing across servers while minimizing application response time. Zhang et al. \cite{11164488} proposed a Dueling Double Deep Q-Network enabled SEIDEL (D3QN-SEIDEL) algorithm for diversified-task co-offloading in Industrial IoT. They constructed a synthetic-expense function considering task correlations, task preferences, and industrial channel characteristics, and categorized production line tasks into three types. The objective is to minimize wireless transmission costs and task execution delay.

\subsection{Cooperative Learning-based Approaches}
Li et al. \cite{10843979} proposed a Multi-Agent Proximal Policy Optimization (MAPPO) algorithm for multi-task scheduling in 6G-enabled intelligent autonomous transport systems. They designed a vehicle-infrastructure network where vehicles can assign multiple computation tasks to edge servers and other idle vehicles, considering vehicle mobility and current workload. The objective is to jointly minimize service request completion time and energy consumption. Chen et al. \cite{10960753} proposed a DRL-based algorithm for vehicular edge computing networks with edge-to-edge collaboration. They employ MAPPO for sub-channel allocation and power control, and PPO for load balancing. The objective is to minimize total task delay. Liu et al. \cite{11039641} proposed the VCPN framework integrating IoTs, vehicles, and edge servers as adaptive computing nodes. They employ the Multi-Agent Deep Deterministic Policy Gradient (MADDPG) algorithm with centralized training and decentralized execution paradigm. The objective is to minimize end-to-end task completion time. Wu et al. \cite{10540320} proposed PG-DDQN, combining Prioritized Experience Replay and Distributed DQN, for proactive caching in cloud-edge computing. They use the multi-agent architecture for centralized training and distributed inference. The objective is to maximize edge hit ratio while minimizing content access latency and traffic cost. Zhang et al. \cite{zhang2024lsia3cs} proposed LsiA3CS, an Asynchronous Advantage Actor-Critic (A3C) based framework for cross-edge cloud collaborative task scheduling in large-scale Industrial IoT. They designed a Markov game model with heuristic policy annealing and action masking to achieve distributed, asynchronous task scheduling across heterogeneous computational resources. The objective is to minimize task completion time and energy consumption. Wang et al. \cite{wang2025tf} proposed TF-DDRL, a Transformer-enhanced Distributed DRL technique based on IMPALA for IoT application scheduling. They integrate Transformers for capturing task dependencies and Prioritized Experience Replay for reducing exploration costs. The objective is to minimize response time, energy consumption, and monetary cost. Raju et al. \cite{10618900} proposed DMITS, using federated DRL with Soft Actor-Critic (SAC) for vehicular fog computing. They model social relationships among vehicles based on bridge centrality and Jaccard similarity. The objective is to maximize task success rate and minimize energy consumption. Chen et al. \cite{10949717} proposed MCM-FDRL, using Federated DRL for vehicular task offloading. Each vehicle acts as an independent agent making offloading decisions using DQN, with model parameters aggregated through federated learning for global optimization. The objective is to minimize task response time while achieving load balancing across edge servers.

\renewcommand{\arraystretch}{1.5}
\begin{table*}[!htbp]
\centering
\caption{Qualitative Comparison of IoT Application Scheduling Approaches}
\label{tab:related_works}
\resizebox{\textwidth}{!}{%
\begin{tabular}{cccccccccccccccccccc}
\hline
\multicolumn{1}{|c|}{\multirow{3}{*}{\textbf{Work}}} & 
\multicolumn{3}{c|}{\textbf{System \& Application Properties}} & 
\multicolumn{4}{c|}{\textbf{Problem Formulation}} & 
\multicolumn{8}{c|}{\textbf{Approach Properties}} & 
\multicolumn{4}{c|}{\textbf{Evaluation}} \\ \cline{2-20}

\multicolumn{1}{|c|}{} & 
\multicolumn{1}{c|}{\multirow{2}{*}{\textbf{\begin{tabular}[c]{@{}c@{}}Task\\Number\end{tabular}}}} & 
\multicolumn{1}{c|}{\multirow{2}{*}{\textbf{\begin{tabular}[c]{@{}c@{}}Task\\Dependency\end{tabular}}}} & 
\multicolumn{1}{c|}{\multirow{2}{*}{\textbf{\begin{tabular}[c]{@{}c@{}}Multi-\\Silo\end{tabular}}}} & 
\multicolumn{1}{c|}{\multirow{2}{*}{\textbf{Time}}} & 
\multicolumn{1}{c|}{\multirow{2}{*}{\textbf{Energy}}} & 
\multicolumn{1}{c|}{\multirow{2}{*}{\textbf{\begin{tabular}[c]{@{}c@{}}Multi-\\Objective\end{tabular}}}} & 
\multicolumn{1}{c|}{\multirow{2}{*}{\textbf{QoS}}} & 
\multicolumn{1}{c|}{\multirow{2}{*}{\textbf{\begin{tabular}[c]{@{}c@{}}Learning\\Paradigm\end{tabular}}}} & 
\multicolumn{1}{c|}{\multirow{2}{*}{\textbf{Algorithm}}} & 
\multicolumn{1}{c|}{\multirow{2}{*}{\textbf{\begin{tabular}[c]{@{}c@{}}Coordination\\Architecture\end{tabular}}}} & 
\multicolumn{1}{c|}{\multirow{2}{*}{\textbf{\begin{tabular}[c]{@{}c@{}}Action\\Space\end{tabular}}}} & 
\multicolumn{1}{c|}{\multirow{2}{*}{\textbf{\begin{tabular}[c]{@{}c@{}}Sparse Reward\\Handling\end{tabular}}}} & 
\multicolumn{1}{c|}{\multirow{2}{*}{\textbf{\begin{tabular}[c]{@{}c@{}}Non-IID\\Handling\end{tabular}}}} & 
\multicolumn{1}{c|}{\multirow{2}{*}{\textbf{Aggregation}}} & 
\multicolumn{1}{c|}{\multirow{2}{*}{\textbf{Robustness}}} & 
\multicolumn{1}{c|}{\multirow{2}{*}{\textbf{Edge}}} &
\multicolumn{1}{c|}{\multirow{2}{*}{\textbf{Cloud}}} &
\multicolumn{1}{c|}{\multirow{2}{*}{\textbf{\begin{tabular}[c]{@{}c@{}}IoT\\Apps\end{tabular}}}} & 
\multicolumn{1}{c|}{\multirow{2}{*}{\textbf{\begin{tabular}[c]{@{}c@{}}Hetero-\\geneity\end{tabular}}}} \\ 
\multicolumn{1}{|c|}{} & 
\multicolumn{1}{c|}{} & 
\multicolumn{1}{c|}{} & 
\multicolumn{1}{c|}{} & 
\multicolumn{1}{c|}{} & 
\multicolumn{1}{c|}{} & 
\multicolumn{1}{c|}{} & 
\multicolumn{1}{c|}{} & 
\multicolumn{1}{c|}{} & 
\multicolumn{1}{c|}{} & 
\multicolumn{1}{c|}{} & 
\multicolumn{1}{c|}{} & 
\multicolumn{1}{c|}{} & 
\multicolumn{1}{c|}{} & 
\multicolumn{1}{c|}{} & 
\multicolumn{1}{c|}{} & 
\multicolumn{1}{c|}{} & 
\multicolumn{1}{c|}{} & 
\multicolumn{1}{c|}{} & 
\multicolumn{1}{c|}{} \\ \hline

\multicolumn{1}{|c|}{\cite{10380323}} & 
\multicolumn{1}{c|}{Multiple} & 
\multicolumn{1}{c|}{\checkmark} & 
\multicolumn{1}{c|}{$\times$} & 
\multicolumn{1}{c|}{\checkmark} & 
\multicolumn{1}{c|}{$\times$} & 
\multicolumn{1}{c|}{$\times$} & 
\multicolumn{1}{c|}{$\times$} & 
\multicolumn{1}{c|}{\multirow{5}{*}{Independent}} & 
\multicolumn{1}{c|}{DQN} & 
\multicolumn{1}{c|}{N/A} & 
\multicolumn{1}{c|}{Fixed} & 
\multicolumn{1}{c|}{$\times$} & 
\multicolumn{1}{c|}{N/A} & 
\multicolumn{1}{c|}{N/A} & 
\multicolumn{1}{c|}{$\times$} & 
\multicolumn{1}{c|}{\Circle} & 
\multicolumn{1}{c|}{\Circle} & 
\multicolumn{1}{c|}{\Circle} & 
\multicolumn{1}{c|}{\checkmark} \\ \cline{1-8} \cline{10-20}

\multicolumn{1}{|c|}{\cite{10989563}} & 
\multicolumn{1}{c|}{Multiple} & 
\multicolumn{1}{c|}{\checkmark} & 
\multicolumn{1}{c|}{$\times$} & 
\multicolumn{1}{c|}{\checkmark} & 
\multicolumn{1}{c|}{\checkmark} & 
\multicolumn{1}{c|}{\checkmark} & 
\multicolumn{1}{c|}{\checkmark} & 
\multicolumn{1}{c|}{} & 
\multicolumn{1}{c|}{DQN} & 
\multicolumn{1}{c|}{N/A} & 
\multicolumn{1}{c|}{Fixed} & 
\multicolumn{1}{c|}{$\times$} & 
\multicolumn{1}{c|}{N/A} & 
\multicolumn{1}{c|}{N/A} & 
\multicolumn{1}{c|}{$\times$} & 
\multicolumn{1}{c|}{\Circle} & 
\multicolumn{1}{c|}{\Circle} & 
\multicolumn{1}{c|}{\LEFTcircle} & 
\multicolumn{1}{c|}{$\times$} \\ \cline{1-8} \cline{10-20}

\multicolumn{1}{|c|}{\cite{10848209}} & 
\multicolumn{1}{c|}{Multiple} & 
\multicolumn{1}{c|}{\checkmark} & 
\multicolumn{1}{c|}{$\times$} & 
\multicolumn{1}{c|}{\checkmark} & 
\multicolumn{1}{c|}{$\times$} & 
\multicolumn{1}{c|}{$\times$} & 
\multicolumn{1}{c|}{$\times$} & 
\multicolumn{1}{c|}{} & 
\multicolumn{1}{c|}{PPO} & 
\multicolumn{1}{c|}{N/A} & 
\multicolumn{1}{c|}{Fixed} & 
\multicolumn{1}{c|}{\checkmark} & 
\multicolumn{1}{c|}{N/A} & 
\multicolumn{1}{c|}{N/A} & 
\multicolumn{1}{c|}{$\times$} & 
\multicolumn{1}{c|}{\Circle} & 
\multicolumn{1}{c|}{\Circle} & 
\multicolumn{1}{c|}{\LEFTcircle} & 
\multicolumn{1}{c|}{\checkmark} \\ \cline{1-8} \cline{10-20}

\multicolumn{1}{|c|}{\cite{wang2024deep}} & 
\multicolumn{1}{c|}{Multiple} & 
\multicolumn{1}{c|}{\checkmark} & 
\multicolumn{1}{c|}{$\times$} & 
\multicolumn{1}{c|}{\checkmark} & 
\multicolumn{1}{c|}{$\times$} & 
\multicolumn{1}{c|}{\checkmark} & 
\multicolumn{1}{c|}{$\times$} & 
\multicolumn{1}{c|}{} & 
\multicolumn{1}{c|}{PPO} & 
\multicolumn{1}{c|}{N/A} & 
\multicolumn{1}{c|}{Fixed} & 
\multicolumn{1}{c|}{\checkmark} & 
\multicolumn{1}{c|}{N/A} & 
\multicolumn{1}{c|}{N/A} & 
\multicolumn{1}{c|}{$\times$} & 
\multicolumn{1}{c|}{\CIRCLE} & 
\multicolumn{1}{c|}{\CIRCLE} & 
\multicolumn{1}{c|}{\CIRCLE} & 
\multicolumn{1}{c|}{\checkmark} \\ \cline{1-8} \cline{10-20}

\multicolumn{1}{|c|}{\cite{11164488}} & 
\multicolumn{1}{c|}{Multiple} & 
\multicolumn{1}{c|}{\checkmark} & 
\multicolumn{1}{c|}{$\times$} & 
\multicolumn{1}{c|}{\checkmark} & 
\multicolumn{1}{c|}{$\times$} & 
\multicolumn{1}{c|}{\checkmark} & 
\multicolumn{1}{c|}{\checkmark} & 
\multicolumn{1}{c|}{} & 
\multicolumn{1}{c|}{D3QN} & 
\multicolumn{1}{c|}{N/A} & 
\multicolumn{1}{c|}{Fixed} & 
\multicolumn{1}{c|}{$\times$} & 
\multicolumn{1}{c|}{N/A} & 
\multicolumn{1}{c|}{N/A} & 
\multicolumn{1}{c|}{$\times$} & 
\multicolumn{1}{c|}{\Circle} & 
\multicolumn{1}{c|}{\Circle} & 
\multicolumn{1}{c|}{\CIRCLE} & 
\multicolumn{1}{c|}{\checkmark} \\ \hline

\multicolumn{1}{|c|}{\cite{10843979}} & 
\multicolumn{1}{c|}{Single} & 
\multicolumn{1}{c|}{$\times$} & 
\multicolumn{1}{c|}{$\times$} & 
\multicolumn{1}{c|}{\checkmark} & 
\multicolumn{1}{c|}{\checkmark} & 
\multicolumn{1}{c|}{\checkmark} & 
\multicolumn{1}{c|}{\checkmark} & 
\multicolumn{1}{c|}{\multirow{8}{*}{Cooperative}} & 
\multicolumn{1}{c|}{MAPPO} & 
\multicolumn{1}{c|}{Centralized} & 
\multicolumn{1}{c|}{Fixed} & 
\multicolumn{1}{c|}{\checkmark} & 
\multicolumn{1}{c|}{N/A} & 
\multicolumn{1}{c|}{N/A} & 
\multicolumn{1}{c|}{$\times$} & 
\multicolumn{1}{c|}{\Circle} & 
\multicolumn{1}{c|}{N/A} & 
\multicolumn{1}{c|}{\Circle} & 
\multicolumn{1}{c|}{$\times$} \\ \cline{1-8} \cline{10-20}

\multicolumn{1}{|c|}{\cite{10960753}} & 
\multicolumn{1}{c|}{Single} & 
\multicolumn{1}{c|}{$\times$} & 
\multicolumn{1}{c|}{$\times$} & 
\multicolumn{1}{c|}{\checkmark} & 
\multicolumn{1}{c|}{$\times$} & 
\multicolumn{1}{c|}{$\times$} & 
\multicolumn{1}{c|}{\checkmark} & 
\multicolumn{1}{c|}{} & 
\multicolumn{1}{c|}{MAPPO} & 
\multicolumn{1}{c|}{Centralized} & 
\multicolumn{1}{c|}{Fixed} & 
\multicolumn{1}{c|}{\checkmark} & 
\multicolumn{1}{c|}{N/A} & 
\multicolumn{1}{c|}{N/A} & 
\multicolumn{1}{c|}{$\times$} & 
\multicolumn{1}{c|}{\Circle} & 
\multicolumn{1}{c|}{N/A} & 
\multicolumn{1}{c|}{\LEFTcircle} & 
\multicolumn{1}{c|}{$\times$} \\ \cline{1-8} \cline{10-20}

\multicolumn{1}{|c|}{\cite{11039641}} & 
\multicolumn{1}{c|}{Single} & 
\multicolumn{1}{c|}{$\times$} & 
\multicolumn{1}{c|}{$\times$} & 
\multicolumn{1}{c|}{\checkmark} & 
\multicolumn{1}{c|}{$\times$} & 
\multicolumn{1}{c|}{$\times$} & 
\multicolumn{1}{c|}{\checkmark} & 
\multicolumn{1}{c|}{} & 
\multicolumn{1}{c|}{MADDPG} & 
\multicolumn{1}{c|}{Centralized} & 
\multicolumn{1}{c|}{Fixed} & 
\multicolumn{1}{c|}{$\times$} & 
\multicolumn{1}{c|}{N/A} & 
\multicolumn{1}{c|}{N/A} & 
\multicolumn{1}{c|}{$\times$} & 
\multicolumn{1}{c|}{\Circle} & 
\multicolumn{1}{c|}{\Circle} & 
\multicolumn{1}{c|}{\Circle} & 
\multicolumn{1}{c|}{\checkmark} \\ \cline{1-8} \cline{10-20}

\multicolumn{1}{|c|}{\cite{10540320}} & 
\multicolumn{1}{c|}{Single} & 
\multicolumn{1}{c|}{$\times$} & 
\multicolumn{1}{c|}{$\times$} & 
\multicolumn{1}{c|}{\checkmark} & 
\multicolumn{1}{c|}{$\times$} & 
\multicolumn{1}{c|}{\checkmark} & 
\multicolumn{1}{c|}{$\times$} & 
\multicolumn{1}{c|}{} & 
\multicolumn{1}{c|}{Distributed DQN} & 
\multicolumn{1}{c|}{Centralized} & 
\multicolumn{1}{c|}{Fixed} & 
\multicolumn{1}{c|}{\checkmark} & 
\multicolumn{1}{c|}{N/A} & 
\multicolumn{1}{c|}{N/A} & 
\multicolumn{1}{c|}{$\times$} & 
\multicolumn{1}{c|}{\Circle} & 
\multicolumn{1}{c|}{\Circle} & 
\multicolumn{1}{c|}{\LEFTcircle} & 
\multicolumn{1}{c|}{$\times$} \\ \cline{1-8} \cline{10-20}

\multicolumn{1}{|c|}{\cite{zhang2024lsia3cs}} & 
\multicolumn{1}{c|}{Single} & 
\multicolumn{1}{c|}{$\times$} & 
\multicolumn{1}{c|}{$\times$} & 
\multicolumn{1}{c|}{\checkmark} & 
\multicolumn{1}{c|}{$\times$} & 
\multicolumn{1}{c|}{\checkmark} & 
\multicolumn{1}{c|}{\checkmark} & 
\multicolumn{1}{c|}{} & 
\multicolumn{1}{c|}{A3C} & 
\multicolumn{1}{c|}{Centralized} & 
\multicolumn{1}{c|}{Fixed} & 
\multicolumn{1}{c|}{\checkmark} & 
\multicolumn{1}{c|}{N/A} & 
\multicolumn{1}{c|}{N/A} & 
\multicolumn{1}{c|}{$\times$} & 
\multicolumn{1}{c|}{\Circle} & 
\multicolumn{1}{c|}{\Circle} & 
\multicolumn{1}{c|}{\LEFTcircle} & 
\multicolumn{1}{c|}{\checkmark} \\ \cline{1-8} \cline{10-20}

\multicolumn{1}{|c|}{\cite{wang2025tf}} & 
\multicolumn{1}{c|}{Multiple} & 
\multicolumn{1}{c|}{\checkmark} & 
\multicolumn{1}{c|}{$\times$} & 
\multicolumn{1}{c|}{\checkmark} & 
\multicolumn{1}{c|}{\checkmark} & 
\multicolumn{1}{c|}{\checkmark} & 
\multicolumn{1}{c|}{$\times$} & 
\multicolumn{1}{c|}{} & 
\multicolumn{1}{c|}{IMPALA} & 
\multicolumn{1}{c|}{Centralized} & 
\multicolumn{1}{c|}{Fixed} & 
\multicolumn{1}{c|}{\checkmark} & 
\multicolumn{1}{c|}{N/A} & 
\multicolumn{1}{c|}{N/A} & 
\multicolumn{1}{c|}{$\times$} & 
\multicolumn{1}{c|}{\CIRCLE} & 
\multicolumn{1}{c|}{\CIRCLE} & 
\multicolumn{1}{c|}{\CIRCLE} & 
\multicolumn{1}{c|}{\checkmark} \\ \cline{1-8} \cline{10-20}

\multicolumn{1}{|c|}{\cite{10618900}} & 
\multicolumn{1}{c|}{Multiple} & 
\multicolumn{1}{c|}{\checkmark} & 
\multicolumn{1}{c|}{$\times$} & 
\multicolumn{1}{c|}{\checkmark} & 
\multicolumn{1}{c|}{\checkmark} & 
\multicolumn{1}{c|}{\checkmark} & 
\multicolumn{1}{c|}{\checkmark} & 
\multicolumn{1}{c|}{} & 
\multicolumn{1}{c|}{Federated SAC} & 
\multicolumn{1}{c|}{Centralized} & 
\multicolumn{1}{c|}{Fixed} & 
\multicolumn{1}{c|}{\checkmark} & 
\multicolumn{1}{c|}{N/A} & 
\multicolumn{1}{c|}{Weighted Aggregation} & 
\multicolumn{1}{c|}{$\times$} & 
\multicolumn{1}{c|}{\LEFTcircle} & 
\multicolumn{1}{c|}{\Circle} & 
\multicolumn{1}{c|}{\LEFTcircle} & 
\multicolumn{1}{c|}{\checkmark} \\ \cline{1-8} \cline{10-20}

\multicolumn{1}{|c|}{\cite{10949717}} & 
\multicolumn{1}{c|}{Single} & 
\multicolumn{1}{c|}{$\times$} & 
\multicolumn{1}{c|}{$\times$} & 
\multicolumn{1}{c|}{\checkmark} & 
\multicolumn{1}{c|}{$\times$} & 
\multicolumn{1}{c|}{$\times$} & 
\multicolumn{1}{c|}{$\times$} & 
\multicolumn{1}{c|}{} & 
\multicolumn{1}{c|}{Federated DQN} & 
\multicolumn{1}{c|}{Centralized} & 
\multicolumn{1}{c|}{Fixed} & 
\multicolumn{1}{c|}{$\times$} & 
\multicolumn{1}{c|}{N/A} & 
\multicolumn{1}{c|}{FedAvg} & 
\multicolumn{1}{c|}{$\times$} & 
\multicolumn{1}{c|}{\Circle} & 
\multicolumn{1}{c|}{N/A} & 
\multicolumn{1}{c|}{\LEFTcircle} & 
\multicolumn{1}{c|}{$\times$} \\ \cline{1-8} \cline{10-20}

\multicolumn{1}{|c|}{\textbf{DeFRiS}} & 
\multicolumn{1}{c|}{Multiple} & 
\multicolumn{1}{c|}{\checkmark} & 
\multicolumn{1}{c|}{\checkmark} & 
\multicolumn{1}{c|}{\checkmark} & 
\multicolumn{1}{c|}{\checkmark} & 
\multicolumn{1}{c|}{\checkmark} & 
\multicolumn{1}{c|}{\checkmark} & 
\multicolumn{1}{c|}{} & 
\multicolumn{1}{c|}{Federated Actor-Critic} & 
\multicolumn{1}{c|}{Decentralized} & 
\multicolumn{1}{c|}{Agnostic} & 
\multicolumn{1}{c|}{\checkmark} & 
\multicolumn{1}{c|}{\begin{tabular}[c]{@{}c@{}}Gradient\\Fingerprint\end{tabular}} & 
\multicolumn{1}{c|}{\begin{tabular}[c]{@{}c@{}}Gradient Fingerprint\\+ Gradient Tracking\end{tabular}} & 
\multicolumn{1}{c|}{\begin{tabular}[c]{@{}c@{}}Gradient Fingerprint\\+ Median Absolute Deviation\end{tabular}} & 
\multicolumn{1}{c|}{\CIRCLE} & 
\multicolumn{1}{c|}{\CIRCLE} & 
\multicolumn{1}{c|}{\CIRCLE} & 
\multicolumn{1}{c|}{\checkmark} \\ \hline
\end{tabular}%
}
\noindent\begin{minipage}{\textwidth}
\vspace{0.1em} 
\footnotesize\raggedright
\textit{Note}: For Edge/Cloud columns: \Circle~indicates simulation, \CIRCLE~indicates real deployment. For IoT Apps column: \Circle~indicates fully simulated applications, \LEFTcircle~indicates applications simulated using real-world datasets, and \CIRCLE~indicates real IoT application deployment.
\end{minipage}
\end{table*}

\subsection{A Qualitative Comparison}
To systematically evaluate existing IoT application scheduling approaches and identify research gaps, we construct a qualitative comparison encompassing multiple indicators across four major categories, as presented in Table~\ref{tab:related_works}. 

\subsubsection{Comparative Analysis Dimensions}
In the system and application properties dimension, we examine task number (single-task or multi-task applications), inter-task dependencies (independent or with precedence-successor constraints), and multi-silo architecture support. The problem formulation dimension covers optimization objective selection, including time, energy consumption, multi-objective trade-offs, and QoS constraints (e.g., deadlines). The approach properties dimension characterizes technical differences from multiple perspectives: learning paradigm (independent or cooperative learning), specific algorithm choice, coordination architecture (centralized or decentralized), action space type (fixed or adaptive), sparse reward handling (whether approaches employ mechanisms to address delayed feedback), while particularly focusing on how federated learning approaches handle Non-IID data, model aggregation strategies, and robustness in adversarial environments. The evaluation dimension assesses deployment realism across edge, cloud, and IoT applications, while heterogeneity indicates whether the evaluation considers environmental heterogeneity (e.g., diverse device capabilities, network conditions).

\subsubsection{Research Gap Identification}
Based on the comprehensive comparison presented in Table~\ref{tab:related_works}, we identify four critical research gaps that are intrinsically interconnected and constitute systematic challenges in IoT application scheduling.

\textbf{Gap 1: Lack of Multi-Silo Architectural Support.} 
Among all 14 studied works, only DeFRiS considers the multi-silo scenario, indicating that existing works universally assume scheduling occurs within a single edge cloud environment, which contradicts real-world cross-organizational cooperation scenarios. More critically, among the 9 cooperative learning approaches, all 8 except DeFRiS adopt centralized coordination architectures, which suffer from data privacy concerns, a single point of failure, and limited scalability in geographically distributed scenarios.

\textbf{Gap 2: Dual Absence of Non-IID Handling and Robustness Mechanisms.} 
The Non-IID Handling and Robustness columns reveal that 13 works neither address Non-IID data nor consider robustness mechanisms. This dual absence is particularly problematic as multi-silo scenarios inevitably introduce data distribution divergence across different geographical locations, while distributed learning environments face threats from malicious or faulty nodes uploading poisoned gradients. Even federated learning works (e.g., weighted aggregation in~\cite{10618900}, FedAvg in~\cite{10949717}) merely employ vanilla aggregation schemes without specialized mechanisms for these critical challenges.

\textbf{Gap 3: Limited Action Space Adaptability.} 
The Action Space column shows that 13 works employ fixed action spaces, with only DeFRiS supporting action space agnosticism. Fixed action spaces assume that scheduling decision options remain static at runtime, failing to exploit task divisibility for fine-grained parallel optimization, adapt to dynamic resource availability changes, or adjust allocation granularity based on heterogeneous device capabilities.

\textbf{Gap 4: Insufficient Real-World Validation.} 
Statistics from the Evaluation dimension reveal pervasive validation insufficiency: except DeFRiS, only 2 works (\cite{wang2024deep,wang2025tf}) conducted real edge/cloud deployment, and only 3 works (\cite{wang2024deep,11164488,wang2025tf}) employed real IoT applications. This validation gap leaves the performance transferability from simulation to real deployment unverified and masks practical challenges (e.g., network latency variance, straggler effects, device failures) in heterogeneous environments.

In summary, these four research gaps are deeply interdependent and form a cascading chain of challenges. Multi-silo scenarios (Gap 1) inherently introduce data heterogeneity and security vulnerabilities across distributed sites, necessitating sophisticated Non-IID handling and robustness mechanisms (Gap 2). These distributed heterogeneous environments exhibit dynamic resource availability and varying device capabilities, demanding adaptive action space strategies (Gap 3). Ultimately, addressing these challenges requires rigorous validation through comprehensive real-world deployment (Gap 4). While existing works have advanced individual aspects, none provide an integrated framework to tackle these interconnected challenges holistically, which constitutes the fundamental motivation for DeFRiS.

\section{Problem Formulation}
\label{problem}
This section presents the mathematical formulation of the silo-cooperative IoT scheduling problem. We establish the system model, define the objectives of response time and energy consumption, and formulate the joint optimization problem subject to operational constraints. Finally, we transform the problem into a distributed Markov Decision Process (MDP) to enable federated DRL-based solutions.

\begin{figure}[t]
    \centering
    \includegraphics[width=0.48\textwidth]{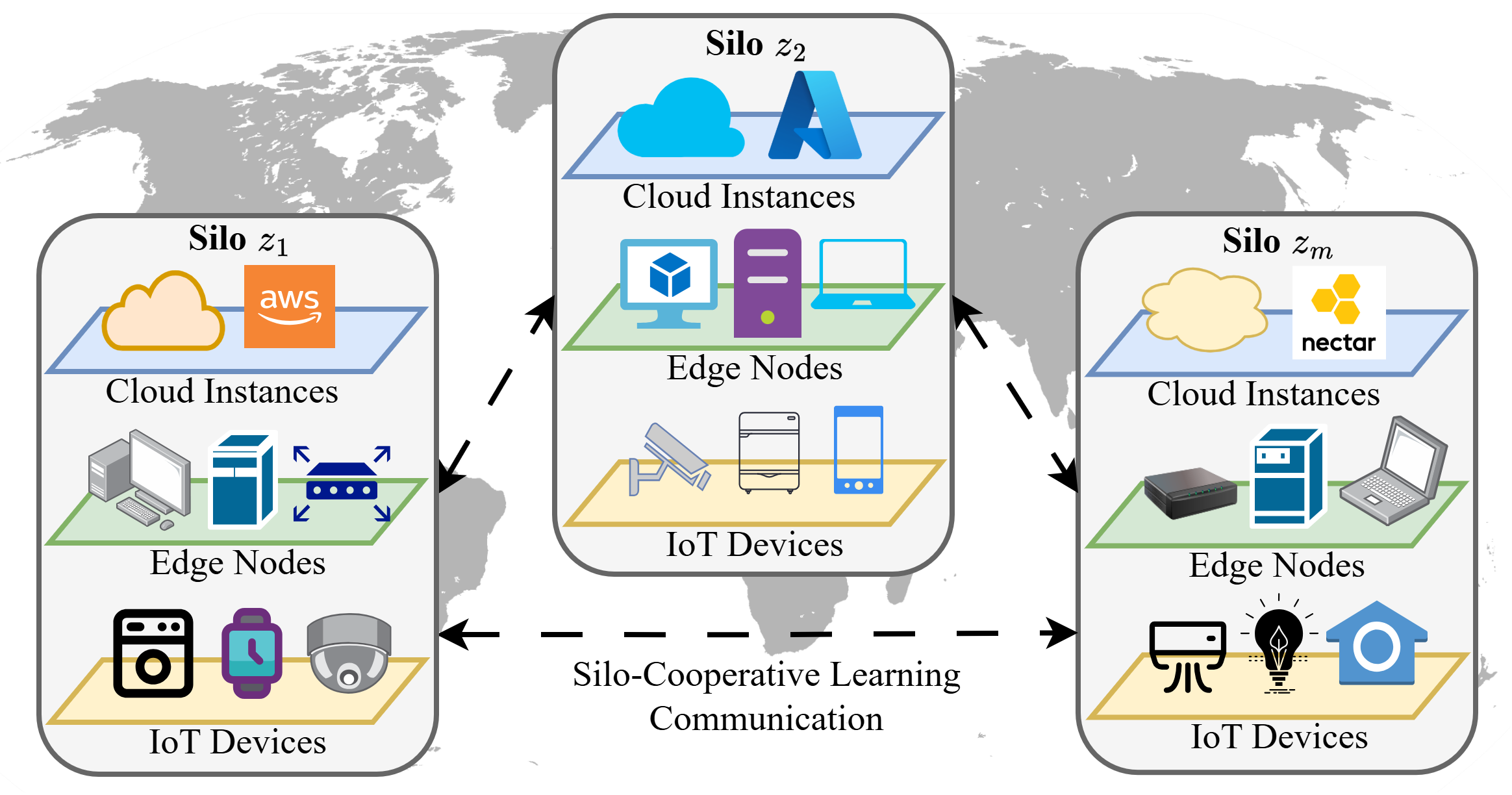}
    \caption{The distributed IoT system model illustrating autonomous silos with heterogeneous resources connected for cooperative learning.}
    \label{fig:system_model}
\end{figure}
\subsection{Silo-Cooperative System Model}
As illustrated in Fig.~\ref{fig:system_model}, we consider a large-scale distributed IoT system comprising multiple dispersed autonomous computing silos $\mathcal{Z} = \{z_1, z_2, \ldots, z_M\}$, each deploying heterogeneous computational resources to support diverse IoT applications. The silos can communicate with each other for cooperative learning purposes while maintaining operational independence. Each autonomous silo $z_i \in \mathcal{Z}$ represents an independent management entity containing a heterogeneous set of computing resources $\mathcal{N}_i$ spanning IoT devices, edge nodes, and cloud instances. 

Each computing entity $n \in \mathcal{N}_i$ is characterized by its computing capability spectrum $\boldsymbol{\rho}_n = (f_n, m_n, d_n, e_n)$, where:
\begin{itemize}
    \item $f_n$ represents the processor frequency (GHz),
    \item $m_n$ represents the available memory capacity (GB),
    \item $d_n$ represents the storage space size (GB),
    \item $e_n$ represents the energy capacity constraint (J), primarily for battery-powered IoT devices; for edge and cloud resources with a stable power supply, $e_n = \infty$.
\end{itemize}

Each silo $z_i$ receives its own set of IoT applications denoted as $\mathcal{T}_i \subseteq \mathcal{T}$, where $\mathcal{T}$ represents the global set of all applications in the system. Considering the complexity and componentized characteristics of modern IoT applications, we model each IoT application $\tau \in \mathcal{T}_i$ as a weighted DAG $\mathcal{G}_\tau = (\mathcal{V}_\tau, \mathcal{E}_\tau, \mathcal{W}_\tau)$, where the vertex set $\mathcal{V}_\tau = \{v_1, v_2, \ldots, v_{|\tau|}\}$ represents atomic tasks, the edge set $\mathcal{E}_\tau \subseteq \mathcal{V}_\tau \times \mathcal{V}_\tau$ encodes data dependency relationships among tasks, and the weight function $\mathcal{W}_\tau: \mathcal{E}_\tau \rightarrow \mathbb{R}_+$ quantifies the data transfer volume between tasks.

We assume that IoT applications provide predefined information upon submission, which is reasonable in practice since IoT applications typically have well-defined computational patterns and explicit QoS requirements. Based on this assumption, each application $\tau$ has an overall deadline constraint $T_\tau^{deadline} \in \mathbb{R}_+$ (in ms), and each task $v_j \in \mathcal{V}_\tau$ is described by the following attribute tuple:
\begin{align}
v_j = \langle f_j^{req}, m_j^{req}, d_j^{req} \rangle,
\end{align}
where:
\begin{itemize}
    \item $f_j^{req}$ is the required CPU cycles (Million cycles),
    \item $m_j^{req}$ is the required memory size (MB),
    \item $d_j^{req}$ is the required storage space (MB).
\end{itemize}

The edge set $\mathcal{E}_\tau$ represents precedence constraints between tasks, where $(v_i, v_j) \in \mathcal{E}_\tau$ indicates that task $v_i$ must complete before task $v_j$ can begin execution. The weight function $\mathcal{W}_\tau: \mathcal{E}_\tau \rightarrow \mathbb{R}_+$ quantifies the data volume (in MB) that must be transferred from task $v_i$ to task $v_j$, formally defined as:

\begin{align}
\mathcal{W}_\tau(v_i, v_j) = \text{DataSize}(v_i \rightarrow v_j), \quad \forall (v_i, v_j) \in \mathcal{E}_\tau.
\end{align}

Each application $\tau \in \mathcal{T}_i$ requires resource allocation decisions for its constituent tasks within silo $z_i$. The scheduling decision for silo $z_i$ is represented by allocation function $\boldsymbol{x}_i: \bigcup_{\tau \in \mathcal{T}_i} \mathcal{V}_\tau \rightarrow \mathcal{N}_i$, where $\boldsymbol{x}_i(v_j)$ denotes the computing entity assigned to task $v_j$ within silo $z_i$.

\subsection{Response Time Objective Function}
IoT applications typically manifest as task graphs with complex dependency relationships, where the total response time is determined by the longest processing path rather than the sum of individual task times. For any processing path $p \in \mathcal{P}(\tau)$, where $\mathcal{P}(\tau)$ denotes the set of all possible execution paths from source to sink tasks in application $\tau$, its total response time consists of three time overheads: computational overhead (actual processing time of tasks on assigned resources), communication overhead (data transmission when dependent tasks are distributed across different computing nodes), and queuing delay caused by resource contention:

\begin{align}
\mathcal{T}_{\text{path}}(p, \boldsymbol{x}_i) &= \sum_{v_j \in p} T_{\text{proc}}(v_j, \boldsymbol{x}_i) + \sum_{(v_j,v_k) \in p} T_{\text{comm}}(v_j, v_k, \boldsymbol{x}_i) \notag \\
&\quad + \sum_{v_j \in p} T_{\text{queue}}(v_j, \boldsymbol{x}_i).
\end{align}

The processing time directly depends on the task's computational requirements and the processing capacity of the assigned resource. For task $v_j$, its processing time on the assigned resource is:
\begin{align}
T_{\text{proc}}(v_j, \boldsymbol{x}_i) = \frac{f_j^{\text{req}}}{f_{\boldsymbol{x}_i(v_j)}}.
\end{align}

When dependent tasks are assigned to different resources, data transmission becomes inevitable. Communication delay comprises two components: transmission time proportional to data volume and fixed network round-trip overhead:
\begin{align}
\scalebox{0.95}{$\displaystyle T_{\text{comm}}(v_j, v_k, \boldsymbol{x}_i) = \frac{\mathcal{W}_\tau(v_j, v_k)}{\text{BW}_{\boldsymbol{x}_i(v_j), \boldsymbol{x}_i(v_k)}} + \text{RTT}_{\boldsymbol{x}_i(v_j), \boldsymbol{x}_i(v_k)}$},
\end{align}
where $\text{BW}_{n_1, n_2}$ represents the available bandwidth between computing entities $n_1$ and $n_2$, and $\text{RTT}_{n_1, n_2}$ is the round-trip time.

In resource-constrained IoT environments, multiple tasks competing for the same computational resource is very common. Queuing delay reflects the impact of such resource contention, calculated as the remaining processing time of all tasks that arrived earlier at the same resource and have not yet completed:
\begin{align}
T_{\text{queue}}(v_j, \boldsymbol{x}_i) = \sum_{\substack{v_l: \boldsymbol{x}_i(v_l) = \boldsymbol{x}_i(v_j) \\ \text{arrive}(v_l) < \text{arrive}(v_j)}} T_{\text{proc}}(v_l, \boldsymbol{x}_i).
\end{align}

With the single-path latency model, we can determine the overall response time of the application. Since different paths in the DAG can be processed in parallel, the application's response time is determined by the longest path (critical path):
\begin{align}
\mathcal{T}_{\text{critical}}(\tau, \boldsymbol{x}_i) = \max_{p \in \mathcal{P}(\tau)} \mathcal{T}_{\text{path}}(p, \boldsymbol{x}_i).
\end{align}

However, optimizing expected response time alone is insufficient. IoT environments exhibit high uncertainty: network condition fluctuations, device performance variations, and randomness in task arrival patterns can all lead to dramatic performance fluctuations. For critical applications, occasional extremely long delays may be more unacceptable than sustained minor delay increases. To simultaneously guarantee average performance and QoS stability, we introduce Conditional Value at Risk (CVaR) to control tail risk. CVaR$_\alpha$ is defined as the conditional expectation beyond the $\alpha$-quantile:
\begin{align}
\text{CVaR}_\alpha[X] = \mathbb{E}[X | X \geq \text{VaR}_\alpha(X)].
\end{align}
For example, CVaR$_{0.95}$ represents the average response time of the slowest 5\% of applications, effectively quantifying system performance under extreme conditions. In practice, we employ the Rockafellar-Uryasev formulation \cite{rockafellar2000optimization} for CVaR computation:
\begin{align}
\text{CVaR}_\alpha[X] = \min_{\eta} \left\{\eta + \frac{1}{1-\alpha} \mathbb{E}\left[(X-\eta)_+\right]\right\},
\end{align}
where $\eta$ is the threshold parameter and $(X-\eta)_+ = \max(0, X-\eta)$ captures excess beyond the threshold. The optimal $\eta^*$ equals the $\alpha$-quantile VaR$_\alpha(X)$. For empirical estimation with batch size $B$, we approximate $\eta^*$ using the sample $\alpha$-quantile $\hat{\eta}$ and compute:
\begin{align}
\widehat{\text{CVaR}}_\alpha = \hat{\eta} + \frac{1}{(1-\alpha)B} \sum_{k=1}^B \left(\mathcal{T}_{\text{critical},k} - \hat{\eta}\right)_+.
\end{align}

For silo $z_i$ processing its application set $\mathcal{T}_i$, the response time objective function is:
\begin{align}
\label{eq:time}
\scalebox{0.85}{$\displaystyle \mathcal{L}_{\text{RT}}^{(i)}(\boldsymbol{x}_i) = \sum_{\tau \in \mathcal{T}_i} \left[ \mathbb{E} \left[ \mathcal{T}_{\text{critical}}(\tau, \boldsymbol{x}_i) \right] + \beta \cdot \text{CVaR}_\alpha\left[\mathcal{T}_{\text{critical}}(\tau, \boldsymbol{x}_i)\right] \right]$},
\end{align}
where the first term optimizes average response time and the second term controls worst-case performance across all applications in silo $z_i$. Parameter $\beta$ balances the two objectives: when $\beta = 0$, focus is on average performance, and as $\beta$ increases, more emphasis is placed on performance stability. This design ensures that scheduling strategies can achieve excellent average performance while providing reliable QoS guarantees for delay-sensitive IoT applications.

\subsection{Energy Consumption Objective Function}
We decompose the total energy consumption into three main components: computation energy, communication energy, and idle maintenance energy. Computation energy is the primary energy consumption source during task execution, directly depending on task computational intensity and resource processing capacity:
\begin{align}
E_{comp}(v_j, \boldsymbol{x}_i) = P_{comp}^{(\boldsymbol{x}_i(v_j))} \cdot T_{proc}(v_j, \boldsymbol{x}_i),
\end{align}
where $P_{comp}^{(n)}$ is the computation power of resource $n$ when executing tasks.

When dependent tasks are assigned to different resources, data transmission generates additional communication energy consumption for both sending and receiving endpoints:
\begin{align}
\scalebox{0.8}{$\displaystyle E_{comm}(v_j, v_k, \boldsymbol{x}_i) = \left[P_{send}^{(\boldsymbol{x}_i(v_j))} + P_{recv}^{(\boldsymbol{x}_i(v_k))}\right] \cdot T_{comm}(v_j, v_k, \boldsymbol{x}_i)$},
\end{align}
where $P_{send}^{(n)}$ represents the transmission power of resource $n$ when sending data, and $P_{recv}^{(n)}$ represents the reception power of resource $n$ when receiving data. Both endpoints consume energy for the entire communication duration $T_{comm}(v_j, v_k, \boldsymbol{x}_i)$ as they are simultaneously engaged in the data transfer process.

Even when no tasks are executing, computing resources still consume energy to maintain their available state:
\begin{align}
E_{idle}(n) = P_{standby}^{(n)} \cdot T_{idle}^{(n)},
\end{align}
where $P_{standby}^{(n)}$ is the standby power of resource $n$, and $T_{idle}^{(n)}$ is the idle time of resource $n$.

Integrating the three components, the total energy consumption of silo $z_i$ under scheduling decision $\boldsymbol{x}_i$ is:
\begin{align}
&\mathcal{L}_{Energy}^{(i)}(\boldsymbol{x}_i) = \sum_{\tau \in \mathcal{T}_i} \left[ \sum_{v_j \in \mathcal{V}_\tau} E_{comp}(v_j, \boldsymbol{x}_i) \right. \notag \\
&\quad \left. + \sum_{(v_j,v_k) \in \mathcal{E}_\tau} E_{comm}(v_j, v_k, \boldsymbol{x}_i) \right] + \sum_{n \in \mathcal{N}_i} E_{idle}(n).
\end{align}

\subsection{Joint Optimization Problem}
Considering that response time and energy consumption have different dimensions and numerical ranges, we first normalize the two objective functions. For a single silo $z_i$, the normalized objective functions are:
\begin{align}
\hat{\mathcal{L}}_{RT}^{(i)}(\boldsymbol{x}_i) = \frac{\mathcal{L}_{RT}^{(i)}(\boldsymbol{x}_i) - \mathcal{L}_{RT,min}^{(i)}}{\mathcal{L}_{RT,max}^{(i)} - \mathcal{L}_{RT,min}^{(i)}},
\end{align}

\begin{align}
\hat{\mathcal{L}}_{Energy}^{(i)}(\boldsymbol{x}_i) = \frac{\mathcal{L}_{Energy}^{(i)}(\boldsymbol{x}_i) - \mathcal{L}_{Energy,min}^{(i)}}{\mathcal{L}_{Energy,max}^{(i)} - \mathcal{L}_{Energy,min}^{(i)}},
\end{align}
where the maximum and minimum values are obtained through historical scheduling data or theoretical analysis. The single-silo optimization problem is formulated as:
\begin{align}
\label{eq:target}
\scalebox{0.8}{$\displaystyle \min_{\boldsymbol{x}_i} \mathcal{L}_{Total}^{(i)}(\boldsymbol{x}_i) = \min_{\boldsymbol{x}_i} \left[ \lambda_{RT}^{(i)} \cdot \hat{\mathcal{L}}_{RT}^{(i)}(\boldsymbol{x}_i) + \lambda_{Energy}^{(i)} \cdot \hat{\mathcal{L}}_{Energy}^{(i)}(\boldsymbol{x}_i) \right]$},
\end{align}
where the weight coefficients satisfy $\lambda_{RT}^{(i)} + \lambda_{Energy}^{(i)} = 1$ and $\lambda_{RT}^{(i)}, \lambda_{Energy}^{(i)} \geq 0$, reflecting silo $i$'s preference between performance and energy efficiency.

From a system-wide perspective, the optimization objective is to minimize the weighted total cost of all silos:
\begin{align}
\label{eq:targets_w}
\min_{\{\boldsymbol{x}_i\}_{i=1}^M} \sum_{i=1}^M \omega_i \cdot \mathcal{L}_{Total}^{(i)}(\boldsymbol{x}_i),
\end{align}
where $\omega_i$ is the importance weight of silo $i$, satisfying $\sum_{i=1}^M \omega_i = 1$. The above optimization problem is subject to the following constraints:

\textbf{Resource capacity constraints:} At any given time, the total resource demands of all concurrently executing tasks on a resource cannot exceed the resource's capacity:
\begin{align}
\label{eq:capacity}
\sum_{\substack{v_j: \boldsymbol{x}_i(v_j) = n \\ t \in [t_{start}(v_j), t_{finish}(v_j)]}} (f_j^{req}, m_j^{req}, &d_j^{req}) \preceq (f_n, m_n, d_n), \notag \\
&\forall t \in \mathbb{R}_+, \forall n \in \mathcal{N}_i,
\end{align}
where $\preceq$ denotes element-wise inequality (i.e., $(a_1, a_2) \preceq (b_1, b_2)$ means $a_1 \leq b_1$ and $a_2 \leq b_2$), and $t_{start}(v_j, \boldsymbol{x}_i)$ and $t_{finish}(v_j, \boldsymbol{x}_i)$ represent the start and finish times of task $v_j$ under allocation $\boldsymbol{x}_i$.

\textbf{Energy budget constraints:} For resources with limited energy capacity (i.e., battery-powered IoT devices), the total energy consumption cannot exceed the available energy budget:
\begin{align}
\label{eq:budget}
&\sum_{v_j: \boldsymbol{x}_i(v_j) = n} E_{comp}(v_j, \boldsymbol{x}_i) + \sum_{(v_j,v_k): \boldsymbol{x}_i(v_j) = n} E_{comm}^{send}(v_j, v_k, \boldsymbol{x}_i) \notag \\
&\quad + \sum_{(v_j,v_k): \boldsymbol{x}_i(v_k) = n} E_{comm}^{recv}(v_j, v_k, \boldsymbol{x}_i) + E_{idle}(n) \leq e_n, \notag \\
&\quad \forall n \in \mathcal{N}_i \text{ with } e_n < \infty.
\end{align}

\textbf{Deadline constraints:} Each application's critical path execution time must not exceed its specified deadline:
\begin{align}
\label{eq:deadline}
\mathcal{T}_{critical}(\tau, \boldsymbol{x}_i) \leq T_\tau^{deadline}, \quad \forall \tau \in \mathcal{T}_i.
\end{align}

\textbf{Precedence constraints:} Tasks with dependencies must execute in the correct order, with predecessor tasks completing and data transmission finishing before their successors can begin:
\begin{align}
\label{eq:precedence}
&t_{finish}(v_j, \boldsymbol{x}_i) + T_{comm}(v_j, v_k, \boldsymbol{x}_i) \leq t_{start}(v_k, \boldsymbol{x}_i), \notag \\
&\quad \forall (v_j, v_k) \in \mathcal{E}_\tau, \tau \in \mathcal{T}_i.
\end{align}

\textbf{Assignment constraints:} Each task must be assigned to exactly one resource within its silo:
\begin{align}
\label{eq:assignment}
\boldsymbol{x}_i(v_j) \in \mathcal{N}_i, \quad \forall v_j \in \bigcup_{\tau \in \mathcal{T}_i} \mathcal{V}_\tau.
\end{align}

This optimization problem exhibits the following characteristics: First is the large-scale state space, where the system state includes resource states, task queues, and network conditions of all silos, with the state space growing exponentially with the number of silos and resource scales. Second is the delayed reward property, where the effects of scheduling decisions are often observed only after task execution completion, forming a typical delayed feedback problem. Third is the distributed coordination challenge, where silos need to cooperate effectively while maintaining autonomy, making traditional centralized optimization approaches difficult to apply.

These characteristics make the problem naturally suitable for MDP modeling, learning optimal scheduling policies through DRL via interaction with the environment. Meanwhile, the coordination requirements in distributed environments provide motivation for applying federated learning approaches.

\subsection{MDP Formulation}
We model the silo-cooperative IoT scheduling problem as a distributed MDP. The entire system can be represented as a tuple:
\begin{align}
\langle \mathcal{S}, \mathcal{A}, P, \mathcal{R}, \gamma \rangle,
\end{align}
where:
\begin{itemize}
    \item $\mathcal{S} = \mathcal{S}_1 \times \mathcal{S}_2 \times \cdots \times \mathcal{S}_M$ is the global state space.
    \item $\mathcal{A} = \mathcal{A}_1 \times \mathcal{A}_2 \times \cdots \times \mathcal{A}_M$ is the joint action space.
    \item $P: \mathcal{S} \times \mathcal{A} \times \mathcal{S} \rightarrow [0,1]$ is the state transition probability function.
    \item $\mathcal{R}: \mathcal{S} \times \mathcal{A} \rightarrow \mathbb{R}$ is the reward function.
    \item $\gamma \in [0,1)$ is the discount factor.
\end{itemize}

\subsubsection{State Space Design}
The local state $s_i^{(t)} \in \mathcal{S}_i$ of each silo $z_i$ contains four key components:
\begin{align}
s_i^{(t)} = \langle \mathbf{U}_i^{(t)}, \mathbf{Q}_i^{(t)}, \mathbf{W}_i^{(t)}, \mathbf{T}_i^{(t)} \rangle,
\end{align}
where $\mathbf{U}_i^{(t)}$ records the resource utilization of computing nodes within the silo, including current occupancy rates of CPU, memory, storage, and energy, providing standardized resource information. $\mathbf{Q}_i^{(t)}$ represents the current task queue lengths of computing nodes, reflecting system load distribution and potential scheduling bottlenecks. $\mathbf{W}_i^{(t)}$ describes the network topology and connectivity within the silo, with diagonal elements recording network bandwidth of nodes and off-diagonal elements recording inter-node communication delays. $\mathbf{T}_i^{(t)}$ contains feature information of pending applications, such as the number of tasks, resource requirements, and deadline constraints.

\subsubsection{Action Space Design}
The action space $\mathcal{A}_i$ of silo $z_i$ corresponds to task-to-resource allocation decisions. For the current task $v_{current}$ to be scheduled, the action is defined as:
\begin{align}
a_i^{(t)} = \boldsymbol{x}_i(v_{current}) \in \mathcal{N}_i,
\end{align}
selecting a computing node within the silo for executing the current task.

We distinguish between hard constraints that must be physically satisfied and soft constraints that can be violated with performance penalties. Hard constraints include resource capacity (Eq. \ref{eq:capacity}), energy budget (Eq. \ref{eq:budget}), precedence (Eq. \ref{eq:precedence}), and assignment constraints (Eq. \ref{eq:assignment}), as violations would result in physically infeasible or logically incorrect executions. Soft constraints include deadline requirements (Eq. \ref{eq:deadline}), where violations lead to QoS degradation but tasks can still be completed.

To ensure physically feasible scheduling decisions, we define the constrained action set that satisfies all hard constraints:
$$\small \mathcal{A}_i^{feasible}(s_i^{(t)}) = \{a_i \in \mathcal{A}_i : \text{satisfying Eqs. \ref{eq:capacity}, \ref{eq:budget}, \ref{eq:precedence}, and \ref{eq:assignment}}\}.$$

\subsubsection{State Transition Function}
The state transition probability function $P(s^{(t+1)} | s^{(t)}, a^{(t)})$ characterizes the system's state evolution after executing joint actions. The transition process includes deterministic factors (such as resource utilization updates and task queue changes) and stochastic factors (such as new application arrivals, task completion time fluctuations, and network condition changes).

\subsubsection{Reward Function Design}
The reward function is directly based on Eq. \ref{eq:target}, using a negative cost formulation:
\begin{align}
r_i^{(t)} = -\mathcal{L}_{Total}^{(i)}(\boldsymbol{x}_i^{(t)}) - C \cdot \mathbb{I}_d,
\end{align}
where $\mathbb{I}_d$ is an indicator function that equals 1 when soft constraints (deadline requirements in Eq. \ref{eq:deadline}) are violated and 0 otherwise. $C$ is a large positive constant that severely penalizes deadline violations to guide policy learning while maintaining decision flexibility.

\subsubsection{Policy and Objective Function}
Each silo $z_i$ maintains a parameterized stochastic policy $\pi_{\theta_i}: \mathcal{S}_i \rightarrow \Delta(\mathcal{A}_i)$, making independent decisions based on local observations:
\begin{align}
\pi_{\theta_i}(a_i | s_i) = P(a_i | s_i; \theta_i).
\end{align}

The global joint policy is defined as the product of local policies of all silos:
\begin{align}
\pi(a_1, \ldots, a_M | s_1, \ldots, s_M) = \prod_{i=1}^M \pi_{\theta_i}(a_i | s_i).
\end{align}

The distributed optimization objective is to maximize the expected cumulative discounted reward across all silos:
\begin{align}
\max_{\{\theta_i\}_{i=1}^M} \sum_{i=1}^M \omega_i \mathbb{E}_{\pi_{\theta_i}} \left[ \sum_{t=0}^{\infty} \gamma^t r_i^{(t)} \right],
\end{align}
where $\omega_i$ is the importance weight of silo $i$ from Eq. \ref{eq:targets_w}, satisfying $\sum_{i=1}^M \omega_i = 1$.

\section{DeFRiS Framework}
\label{framework}
Silo-cooperative IoT scheduling poses three fundamental challenges for standard DRL: \textbf{(i) heterogeneous action spaces}: silos expose resource sets with different cardinalities and semantics, preventing straightforward parameter sharing and transfer across policies; \textbf{(ii) stable and sample-efficient local optimization}: each silo must learn from limited, privacy-constrained experience under sparse and delayed rewards; updates must remain stable and keep models ready for subsequent federation; \textbf{(iii) Non-IID aggregation bias and attacks}: workload distributions differ significantly across silos, and malicious participants may inject corrupted parameters, making robust aggregation essential for effective cooperation.

\begin{figure*}[t]
    \centering
    \includegraphics[width=\textwidth]{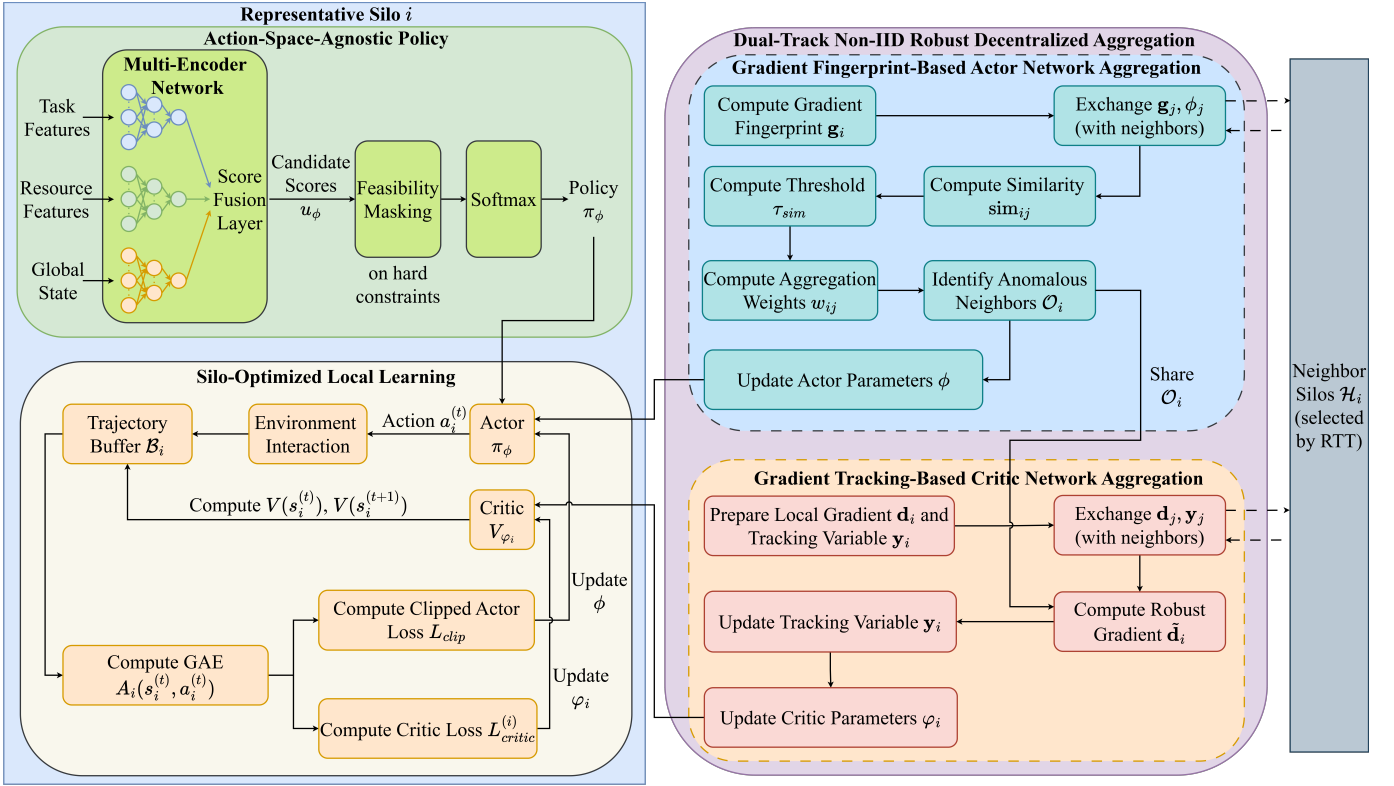}
    \caption{The overview of the DeFRiS framework. It consists of three synergistic components: (1) Action-Space-Agnostic Policy enables parameter sharing across heterogeneous silos via candidate scoring; (2) Silo-Optimized Local Learning ensures stable convergence under sparse rewards using GAE and clipped updates; (3) Dual-Track Non-IID Robust Decentralized Aggregation facilitates robust and similarity-aware knowledge transfer through gradient fingerprints and tracking.}
    \label{fig:defris_framework}
\end{figure*}

As depicted in Fig.~\ref{fig:defris_framework}, DeFRiS addresses these challenges with three complementary components: \textbf{Action-Space-Agnostic Policy} that models resource selection via candidate scoring with masking, enabling cross-silo parameter sharing; \textbf{Silo-Optimized Local Learning} that uses advantage estimation and clipped update steps to provide stable on-policy improvement; and \textbf{Dual-Track Non-IID Robust Decentralized Aggregation}, employing gradient fingerprint-based similarity assessment and gradient tracking mechanism to mitigate aggregation bias while detecting and filtering anomalous behavior.

\subsection{Action-Space-Agnostic Policy}
\label{asap}
Different silos possess computing resource collections of varying scales and configurations, leading to significant differences in action space dimensions. For example, an edge-computing-dominant silo might contain 20 edge nodes and 5 cloud instances, while a cloud-computing-dominant silo might contain 8 edge nodes and 15 cloud instances. Traditional policy networks require output layer weight matrices that must match the action space size, preventing parameter sharing when different silos have different action space dimensions.

\subsubsection{Candidate Scoring Design Philosophy}
To solve the heterogeneous action space problem, we redesign the policy function $\pi_{\theta_i}(a_i | s_i)$ defined in the MDP into a candidate scoring-based form. The core idea is to design a scoring network that computes suitability scores for each candidate resource node, then obtains policy probability distributions through normalization.

Specifically, we redefine the policy function as:
\begin{align}
\pi_\phi(n|s_i) = \frac{\exp(u_\phi(s_i, n))}{\sum_{n' \in \mathcal{A}_i^{feasible}(s_i)} \exp(u_\phi(s_i, n'))},
\end{align}
where $u_\phi(s_i, n)$ is the state-resource scoring function, $n \in \mathcal{N}_i$ represents candidate resource nodes, and $\mathcal{A}_i^{feasible}(s_i)$ is the feasible action set satisfying hard constraints. The key innovation is the independence of the parameters $\phi$ from the resource count, enabling the same scoring function $u_\phi$ to handle candidate sets of arbitrary scale.

This remodeling brings three important advantages. First, parameter shareability allows all silos to use the same $\phi$ parameters because the scoring function only needs to process individual state-resource pairs. Second, scalability ensures that when a silo's resource configuration changes, no model retraining is needed. Third, semantic consistency enables the scoring function to learn general matching patterns between states and resource features, and this knowledge has good transferability across different silos.

\subsubsection{Multi-Encoder Network Architecture Design}
The scoring function $u_\phi(s_i, n)$ adopts a multi-encoder architecture with multiple fully connected layers that separately encode task features, resource features, and global state:

\begin{itemize}
\item \textbf{Task Encoder} processes features of the current task to be scheduled:
\begin{align}
\mathbf{h}_{task} = \text{ReLU}(\mathbf{W}_{task} \cdot \psi_{task}(s_i) + \mathbf{b}_{task}),
\end{align}
where $\psi_{task}(s_i)$ extracts features of the current task such as requirements, deadline, etc.

\item \textbf{Resource Encoder} processes features of candidate resource nodes:
\begin{align}
\scalebox{0.85}{$\displaystyle \mathbf{h}_{resource} = \text{ReLU}(\mathbf{W}_{resource} \cdot \psi_{resource}(n) + \mathbf{b}_{resource})$},
\end{align}
where $\psi_{resource}(n)$ includes features of resource $n$ such as index, capacity, current load, queue length, network conditions, etc.

\item \textbf{Global State Encoder} processes system-wide state information:
\begin{align}
\mathbf{h}_{global} = \text{ReLU}(\mathbf{W}_{global} \cdot \psi_{global}(s_i) + \mathbf{b}_{global}),
\end{align}
where $\psi_{global}(s_i)$ contains global information such as overall load distribution across all resources, network topology, etc.

\item \textbf{Score Fusion Layer} computes final task-resource matching scores:
\begin{align}
u_\phi(s_i, n) &= \mathbf{w}^T \cdot \text{ReLU}([\mathbf{h}_{task}; \mathbf{h}_{resource}; \mathbf{h}_{global}; \notag \\ 
&\quad \mathbf{h}_{task} \odot \mathbf{h}_{resource}]),
\end{align}
where the concatenation operation $[\mathbf{h}_{task}; \mathbf{h}_{resource}; \mathbf{h}_{global}]$ captures independent features of tasks, resources, and global state, while the element-wise product $\mathbf{h}_{task} \odot \mathbf{h}_{resource}$ models interactive matching effects between tasks and resources.
\end{itemize}

\subsubsection{Feasibility Masking}
We handle hard constraints through a feasibility masking mechanism. The policy function can be equivalently expressed as:
\begin{align}
\pi_\phi(n|s_i) &= \frac{\exp(u_\phi(s_i, n))}{\sum_{n' \in \mathcal{A}_i^{feasible}(s_i)} \exp(u_\phi(s_i, n'))} \notag \\
&= \frac{\exp(u_\phi(s_i, n)) \cdot \mathbb{I}_i(s_i,n)}{\sum_{n' \in \mathcal{N}_i} \exp(u_\phi(s_i, n')) \cdot \mathbb{I}_i(s_i,n')},
\end{align}
where $\mathbb{I}_i(s_i,n) \in \{0,1\}$ is the feasibility mask for silo $i$. When resource $n$ satisfies all hard constraints for the current task in silo $i$, $\mathbb{I}_i(s_i,n) = 1$, otherwise it equals 0. This design ensures that the policy only computes probability distributions over physically feasible action spaces while maintaining the mathematical properties of softmax normalization.

\subsection{Silo-Optimized Local Learning}
Silo-cooperative IoT scheduling faces the challenge of learning effective policies from sparse, delayed rewards. To address this, we develop a specialized local learning mechanism that combines Actor-Critic architecture with advanced training techniques to ensure both sample efficiency and stability for subsequent federated aggregation.

\subsubsection{Actor-Critic Architecture}
We adopt an Actor-Critic architecture that leverages the action-space-agnostic policy while adding a dedicated value network for stable learning. In the following discussion, we adopt time-indexed notation $(\cdot)^{(t)}$ to support the temporal learning dynamics.

The Actor network directly uses the policy $\pi_\phi(n|s_i^{(t)})$ from the Section \ref{asap} with its multi-encoder scoring function, enabling cross-silo parameter sharing:
\begin{align}
\pi_\phi(n|s_i^{(t)}) = \frac{\exp(u_\phi(s_i^{(t)}, n)) \cdot \mathbb{I}_i(s_i^{(t)},n)}{\sum_{n' \in \mathcal{N}_i} \exp(u_\phi(s_i^{(t)}, n')) \cdot \mathbb{I}_i(s_i^{(t)},n')}.
\end{align}

The Critic network employs independent parameters $\varphi_i$, specifically learning state value functions for silo $i$'s local environment:
\begin{align}
\scalebox{0.9}{$\displaystyle V_{\varphi_i}(s_i^{(t)}) = \mathbf{w}_{val}^T \cdot \text{ReLU}(\mathbf{W}_{val} \cdot \psi_{state}(s_i^{(t)}) + \mathbf{b}_{val}) + b_{val}$},
\end{align}
where $\psi_{state}(s_i^{(t)})$ converts MDP states into unified feature representations, $\mathbf{b}_{val}$ is the hidden layer bias vector, and $b_{val}$ is the output layer bias scalar.

\subsubsection{Stable Local Training}
In IoT scheduling environments, reward signals exhibit typical sparsity and delay characteristics, where the effects of a scheduling decision may only be observed upon completion of the entire application execution. This creates a significant credit assignment problem \cite{sutton1984temporal}, as it becomes difficult to determine which specific scheduling decisions among a long sequence of actions contribute to the final application performance. To address these challenges and ensure training stability, we employ two specialized training techniques:

\begin{itemize}
\item \textbf{Generalized Advantage Estimation (GAE) Mechanism}: To solve the sparse reward problem, we employ GAE to convert sparse terminal rewards into dense intermediate learning signals. GAE combines multi-step temporal difference errors through exponential weighting:
\begin{align}
\label{eq:gae}
A_i(s_i^{(t)},a_i^{(t)}) = \sum_{l=0}^{\infty} (\gamma \lambda)^l \delta_i^{(t+l)},
\end{align}
where $\delta_i^{(t)} = r_i^{(t)} + \gamma V_{\varphi_i}(s_i^{(t+1)}) - V_{\varphi_i}(s_i^{(t)})$ is the Temporal Difference (TD) error for silo $i$ at time $t$. The parameter $\lambda \in [0,1]$ controls the estimation time span: when $\lambda$ approaches 0, it focuses on immediate rewards with low bias but high variance; when $\lambda$ approaches 1, it considers long-term returns with unbiased but high variance estimates. This design enables each intermediate decision to receive gradient signals from final performance, effectively addressing the credit assignment challenge in sparse reward environments.

\item \textbf{Clipped Update Stability Guarantee}: Policy network updates may lead to significant differences between new and old policies, easily triggering training instability in complex scheduling environments. We monitor policy change magnitude through importance sampling ratios:
\begin{align}
\label{eq:imp}
\kappa_i(s_i^{(t)},a_i^{(t)}) = \frac{\pi_\phi(a_i^{(t)}|s_i^{(t)})}{\pi_\phi^{old}(a_i^{(t)}|s_i^{(t)})}.
\end{align}
When $\kappa_i(s_i^{(t)},a_i^{(t)})$ deviates far from 1, it indicates large policy differences. Such dramatic policy changes can cause the agent to take actions that are poorly understood by the current value function, leading to unstable learning or even performance collapse. To prevent this, we employ a clipped objective function that constrains the magnitude of policy updates:
\begin{align}
\label{eq:clip}
&L_{clip}^{(i)}(\phi) = \min\left(\kappa_i(s_i^{(t)},a_i^{(t)}) \cdot A_i(s_i^{(t)},a_i^{(t)}), \right. \notag \\
&\left. \text{clip}(\kappa_i(s_i^{(t)},a_i^{(t)}), 1-\epsilon, 1+\epsilon) \cdot A_i(s_i^{(t)},a_i^{(t)})\right).
\end{align}
The clipping function restricts ratios to the interval $[1-\epsilon, 1+\epsilon]$, ensuring policy update step sizes are controlled within safe ranges and avoiding performance degradation due to single large updates.
\end{itemize}

Algorithm~\ref{alg:silo_local_training} presents the complete silo-optimized local learning procedure. During the experience collection phase (Lines 4-9), each silo executes its current policy $\pi_{\phi}(n|s_i^{(t)})$ with feasibility masking $\mathbb{I}_i(s_i^{(t)},n)$ to gather trajectory data while ensuring all sampled actions satisfy hard constraints, forming the training batch $\mathcal{B}_i$. In the GAE advantage estimation phase (Lines 10-16), the algorithm implements Eq. \ref{eq:gae} through backward recursive computation. The Actor update phase (Lines 17-22) employs the clipped objective as designed in Eq. \ref{eq:clip}, where importance sampling ratios are first computed following Eq. \ref{eq:imp}, then incorporated into the clipped gradient computation:
\begin{align}
\nabla_{\phi} &L_{clip}^{(i)} = \frac{1}{|\mathcal{B}_i|} \sum_{(s_i^{(t)},a_i^{(t)}) \in \mathcal{B}_i} \notag \\
&\ \nabla_{\phi} \min\left(\kappa_i(s_i^{(t)},a_i^{(t)}) \cdot A_i(s_i^{(t)},a_i^{(t)}), \right. \notag \\
&\left. \text{clip}(\kappa_i(s_i^{(t)},a_i^{(t)}), 1-\epsilon, 1+\epsilon) \cdot A_i(s_i^{(t)},a_i^{(t)})\right).
\end{align}
Actor parameter updates follow gradient ascent: $\phi \leftarrow \phi + \alpha_{\phi} \nabla_{\phi} L_{clip}^{(i)}$. The Critic update phase (Lines 23-26) optimizes value prediction accuracy by minimizing the temporal difference loss function:
\begin{align}
L_{critic}^{(i)} = \frac{1}{|\mathcal{B}_i|} \sum_{s_i^{(t)} \in \mathcal{B}_i} \frac{1}{2}(V_{\varphi_i}(s_i^{(t)}) - V_{target}(s_i^{(t)}))^2,
\end{align}
where target values $V_{target}(s_i^{(t)}) = r_i^{(t)} + \gamma V_{\varphi_i}(s_i^{(t+1)})$ incorporate both immediate scheduling rewards and discounted future value estimates. Critic parameter updates follow gradient descent: $\varphi_i \leftarrow \varphi_i - \alpha_{\varphi} \nabla_{\varphi_i} L_{critic}^{(i)}$. 

\begin{algorithm}[htbp]
\footnotesize
\caption{Silo-Optimized Local Learning}
\label{alg:silo_local_training}
\begin{algorithmic}[1]
\REQUIRE Silo $z_i$, initial parameters $\phi^{(0)}, \varphi_i^{(0)}$, learning rates $\alpha_{\phi}, \alpha_{\varphi}$, clipping parameter $\epsilon$, GAE parameter $\lambda$, discount factor $\gamma$
\ENSURE Trained parameters $\phi, \varphi_i$

\FOR{training episode $e = 1, 2, \ldots$}
    \STATE Initialize trajectory buffer $\mathcal{B}_i = \{\}$
    
    \STATE // Experience Collection Phase
    \FOR{step $t = 1, 2, \ldots, T$}
        \STATE Observe current state $s_i^{(t)}$ from local environment
        \STATE Sample action: $a_i^{(t)} \sim \pi_{\phi}(n|s_i^{(t)})$ using feasibility masking $\mathbb{I}_i(s_i^{(t)},n)$
        \STATE Execute action $a_i^{(t)}$, observe reward $r_i^{(t)}$ and next state $s_i^{(t+1)}$
        \STATE Store transition $(s_i^{(t)}, a_i^{(t)}, r_i^{(t)}, s_i^{(t+1)})$ in $\mathcal{B}_i$
    \ENDFOR
    
    \STATE // GAE Advantage Estimation Phase
    \STATE Initialize advantage accumulator $A = 0$
    \FOR{$(s_i^{(t)}, a_i^{(t)}, r_i^{(t)}, s_i^{(t+1)}) \in \mathcal{B}_i$ \textbf{in reverse chronological order}}
        \STATE Compute TD error: $\delta_i^{(t)} = r_i^{(t)} + \gamma V_{\varphi_i}(s_i^{(t+1)}) - V_{\varphi_i}(s_i^{(t)})$
        \STATE Update accumulator: $A = \delta_i^{(t)} + \gamma \lambda A$
        \STATE Set advantage: $A_i(s_i^{(t)}, a_i^{(t)}) = A$
    \ENDFOR
    
    \STATE // Actor Update Phase with Stability Control
    \STATE Store current policy parameters: $\phi^{old} = \phi$
    \STATE Compute importance sampling ratios: $\kappa_i(s_i^{(t)},a_i^{(t)}) = \frac{\pi_{\phi}(a_i^{(t)}|s_i^{(t)})}{\pi_{\phi^{old}}(a_i^{(t)}|s_i^{(t)})}, \forall (s_i^{(t)},a_i^{(t)}) \in \mathcal{B}_i$
    \STATE Compute clipped policy objective: 
    \STATE \quad $L_{clip}^{(i)} = \frac{1}{|\mathcal{B}_i|} \sum_{(s_i^{(t)},a_i^{(t)}) \in \mathcal{B}_i} \min(\kappa_i(s_i^{(t)},a_i^{(t)}) \cdot A_i(s_i^{(t)},a_i^{(t)}), \text{clip}(\kappa_i(s_i^{(t)},a_i^{(t)}), 1-\epsilon, 1+\epsilon) \cdot A_i(s_i^{(t)},a_i^{(t)}))$
    \STATE Update Actor parameters: $\phi \leftarrow \phi + \alpha_{\phi} \nabla_{\phi} L_{clip}^{(i)}$
    
    \STATE // Critic Update Phase
    \STATE Compute value targets: $V_{target}(s_i^{(t)}) = r_i^{(t)} + \gamma V_{\varphi_i}(s_i^{(t+1)}), \forall (s_i^{(t)},a_i^{(t)},r_i^{(t)},s_i^{(t+1)}) \in \mathcal{B}_i$
    \STATE Compute critic loss: $L_{critic}^{(i)} = \frac{1}{|\mathcal{B}_i|} \sum_{s_i^{(t)} \in \mathcal{B}_i} \frac{1}{2}(V_{\varphi_i}(s_i^{(t)}) - V_{target}(s_i^{(t)}))^2$
    \STATE Update Critic parameters: $\varphi_i \leftarrow \varphi_i - \alpha_{\varphi} \nabla_{\varphi_i} L_{critic}^{(i)}$
    
\ENDFOR
\end{algorithmic}
\end{algorithm}

\subsection{Dual-Track Non-IID Robust Decentralized Aggregation}
In distributed IoT scheduling environments, silos face significantly different task workloads and resource configurations, creating severe Non-IID challenges. Additionally, distributed environments may contain faulty nodes, network attacks, or malicious behavior, requiring robust aggregation mechanisms. This section proposes a decentralized robust aggregation mechanism that achieves effective cross-silo knowledge transfer through similarity-aware, anomaly detection, and gradient tracking techniques, while resisting malicious nodes.

\subsubsection{Decentralized Communication and Aggregation Protocol}
To achieve privacy protection and fault tolerance, we adopt a fully decentralized peer-to-peer aggregation architecture. The system constructs a time-varying overlay network $\mathcal{G}_{comm}(t) = (\mathcal{Z}, \mathcal{E}_{comm}(t))$, where the edge set $\mathcal{E}_{comm}(t)$ adjusts dynamically based on network conditions.

Each silo $z_i$ randomly samples $k_{sample}$ candidate silos for Round-Trip Time (RTT) measurement during initialization, where $k_{sample} < M$. Based on the delay measurement results, each silo $z_i$ selects the $d_{max}$ silos with the lowest latency as neighbors, forming the neighbor set $\mathcal{H}_i(t)$. Therefore, each silo maintains a neighbor set $\mathcal{H}_i(t) \subseteq \mathcal{Z} \setminus \{z_i\}$ satisfying the degree constraint $|\mathcal{H}_i(t)| \leq d_{max}$, ensuring communication efficiency and fault tolerance.

Parameter propagation follows a gossip protocol. In each communication round, silo $z_i$ randomly selects partial neighbors from $\mathcal{H}_i(t)$ for parameter exchange. During parameter exchange, neighbor $z_j$ attaches summary information of its current neighbor list, including neighbor identifiers and corresponding RTT values. Silo $z_i$ identifies potential better neighbor candidates based on this recommendation information, selecting those candidates whose RTT is significantly lower than the current worst neighbor for connectivity testing. If the actual RTT of a new candidate is indeed better, it replaces the neighbor with the highest RTT in the current neighbor set, achieving gradual network topology optimization.

Recognizing the different characteristics of the Actor-Critic architecture, we design a dual-track aggregation strategy with periodic aggregation (every $\Omega$ local training rounds). The Actor network generates generalizable scheduling patterns and leverages cross-silo policy knowledge; the Critic network evaluates local state values and needs to maintain accurate modeling of the local environment, thus adopting gradient tracking to preserve personalization.

\subsubsection{Gradient Fingerprint-Based Actor Network Aggregation}
The Actor network needs to learn generalizable scheduling policies, but different silos have varying contribution values. Directly averaging all neighbor parameters may introduce knowledge unsuitable for the local environment, leading to performance degradation. Meanwhile, in open distributed environments, some silos may send abnormal parameters due to hardware failures, network attacks, or malicious behavior, further threatening aggregation effectiveness. Therefore, we design a robust aggregation mechanism based on similarity awareness and anomaly detection that can prioritize learning policy knowledge from silos with similar environments while resisting attacks.

We propose a gradient fingerprint-based similarity measurement to capture the strategic preference patterns of different silos. The fundamental insight is that silos learning effective resource selection policies will exhibit consistent gradient patterns with respect to resource features, reflecting their learned preferences for specific resource attributes. Silos demonstrating similar preference patterns in their optimization dynamics are more likely to benefit from mutual policy knowledge transfer.

During each backpropagation step, the gradient of the candidate scoring function $u_\phi(s_t, n_t)$ with respect to the resource feature encoding $\psi_{\mathrm{resource}}(n_t)$ reveals the model's sensitivity and preference intensity toward different resource attributes. We construct a gradient fingerprint by averaging the resource feature gradients over the most recent $K$ training samples:
\begin{align}
\mathbf{g}_i = \frac{1}{K}\sum_{t=1}^K \frac{\partial u_\phi(s_t,n_t)}{\partial \psi_{\mathrm{resource}}(n_t)}.
\end{align}
Each component of the gradient fingerprint vector $\mathbf{g}_i \in \mathbb{R}^{d_{\mathrm{resource}}}$ corresponds to a resource feature dimension, with its magnitude indicating the strategic importance and optimization direction that silo $i$'s policy assigns to that particular feature. This representation exhibits three properties desirable for heterogeneous federated learning: \emph{(i) cross-silo comparability}, since $\mathbf{g}_i\in\mathbb{R}^{d_{\mathrm{resource}}}$ is defined in the shared resource-feature space $\psi_{\mathrm{resource}}(\cdot)$ and is thus comparable across silos regardless of candidate-set size or identity; \emph{(ii) negligible marginal cost}, because the required gradients are produced by standard backpropagation and averaged locally, with no extra forward passes or cross-silo data exchange; and \emph{(iii) temporal adaptivity}, as maintaining $\mathbf{g}_i$ as a sliding-window average over recent updates makes it responsive to policy evolution and workload shifts, providing a low-latency summary of emerging preference patterns.

Inter-silo similarity is computed using cosine similarity to measure gradient direction alignment:

\begin{align}
\mathrm{sim}_{ij} = \frac{\mathbf{g}_i^\top \mathbf{g}_j}{\|\mathbf{g}_i\| \|\mathbf{g}_j\|}.
\end{align}
Cosine similarity captures gradient direction consistency while maintaining invariance to magnitude differences, ensuring that similarity measurement is unaffected by variations in training progress or data scale across silos. A high similarity score indicates that both silos have developed consistent preferences regarding resource attribute importance, suggesting compatible optimization objectives.

To identify suspicious neighbors, we also employ the gradient fingerprint for anomaly detection. In distributed environments, normally cooperating silos should exhibit consistent resource preference patterns, while abnormal silos (whether due to failures or malicious behavior) may demonstrate dramatically different optimization objectives that manifest as divergent gradient fingerprints. We identify anomalous neighbors based on similarity thresholds $\tau_{\mathrm{sim}}$:
\begin{align}
\mathcal{O}_i = \{j \in \mathcal{H}_i : \mathrm{sim}_{ij} < \tau_{\mathrm{sim}}\},
\end{align}
where $\tau_{\mathrm{sim}}$ is set using robust statistics to account for the non-Gaussian distribution of cosine similarity scores:
\begin{align}
\tau_{\mathrm{sim}} = \text{median}(&\{\mathrm{sim}_{ij} : j \in \mathcal{H}_i\}) - \notag \\
&\xi \cdot \text{MAD}(\{\mathrm{sim}_{ij} : j \in \mathcal{H}_i\}),
\end{align}
where $\text{MAD}$ is the Median Absolute Deviation:
\begin{align}
\text{MAD}&(\{\mathrm{sim}_{ij} : j \in \mathcal{H}_i\}) = \text{median}(\{|\mathrm{sim}_{ij} \notag \\
& - \text{median}(\{\mathrm{sim}_{ik} : k \in \mathcal{H}_i\})| : j \in \mathcal{H}_i\}),
\end{align}
and $\xi$ is a robustness parameter. This robust approach provides reliable anomaly detection for bounded similarity measures without distributional assumptions. 

Aggregation weights are computed through softmax normalization of similarity scores for non-anomalous neighbors:
\begin{align}
w_{ij} = \frac{\exp(\mathrm{sim}_{ij}/\nu)}{\sum_{j' \in \mathcal{H}_i \setminus \mathcal{O}_i} \exp(\mathrm{sim}_{ij'}/\nu)},
\end{align}
where $\nu > 0$ is a temperature parameter controlling the concentration of the similarity distribution.

Actor parameter updates employ robust weighted aggregation. First, anomalous neighbors are excluded, then weighted aggregation is performed on the remaining neighbors' parameters:
\begin{align}
\phi^{(k+1)} = (1-\alpha_{\mathrm{agg}}) \cdot \phi^{(k)} + \alpha_{\mathrm{agg}} \cdot \sum_{j \in \mathcal{H}_i \setminus \mathcal{O}_i} w_{ij} \cdot \phi_j^{(k)},
\end{align}
where $\alpha_{\mathrm{agg}} \in (0,1)$ is the aggregation rate for the Actor network. 

\subsubsection{Gradient Tracking-Based Critic Network Aggregation}
The core function of the Critic network $V_{\varphi_i}(s_i)$ is to accurately evaluate the local state values of silo $z_i$. Direct parameter averaging would destroy the accuracy of value functions because different silos have different state value distributions. To address this challenge, we propose a gradient tracking mechanism that enables silos to maintain personalized value functions while benefiting from neighbors' optimization directions. 

Each silo $z_i$ maintains two key variables: the local gradient $\mathbf{d}_i^{(k)} = \nabla_{\varphi_i} L_{critic}^{(i)}(\varphi_i^{(k)})$ reflecting the local optimization direction under current parameters, and the tracking variable $\mathbf{y}_i^{(k)}$ accumulating gradient information from distributed cooperation. To resist abnormal gradients, we use the same anomalous neighbor set $\mathcal{O}_i$. Each silo computes its own robust gradient by averaging filtered neighbor gradients:
\begin{align}
\tilde{\mathbf{d}}_i^{(k)} = \frac{1}{|\mathcal{H}_i \setminus \mathcal{O}_i| + 1} \sum_{j \in (\mathcal{H}_i \setminus \mathcal{O}_i) \cup \{i\}} \mathbf{d}_j^{(k)}.
\end{align}

The update of tracking variables $\mathbf{y}_i^{(k)}$ follows:
\begin{align}
\mathbf{y}_i^{(k+1)} = \sum_{j \in \mathcal{H}_i \cup \{i\}} c_{ij} \left(\mathbf{y}_j^{(k)} + \tilde{\mathbf{d}}_j^{(k+1)} - \tilde{\mathbf{d}}_j^{(k)}\right).
\end{align}

This formula design has three key components. First are the uniform mixing weights $c_{ij}$, which are chosen to satisfy the row stochastic condition $\sum_{j=1}^M c_{ij} = 1$. Specifically, the weights are defined based on the network topology as $c_{ij} = \frac{1}{|\mathcal{H}_i| + 1}$ for $j \in \mathcal{H}_i \cup \{i\}$ and $c_{ij} = 0$ otherwise. This choice ensures balanced information propagation and convergence to a network-wide consensus while maintaining computational simplicity. Second, the tracking variable $\mathbf{y}_j^{(k)}$ carries historical cooperative gradient information from neighbor silos, enabling cross-silo optimization experience transfer. Most critically, the gradient difference term $\tilde{\mathbf{d}}_j^{(k+1)} - \tilde{\mathbf{d}}_j^{(k)}$ captures the robust gradient change of neighbors from round $k$ to $k+1$, reflecting new optimization directions discovered by neighbors or adaptive adjustments to environmental changes. This design enables each silo to perceive neighbors' optimization dynamics: when neighbors discover better optimization directions, the gradient difference term transmits this improvement information; when environmental changes cause gradient adjustments, other silos can promptly perceive and adjust accordingly. Through mixing and accumulation of tracking variables, the optimization wisdom of the entire network is shared while each silo maintains the ability to adapt to local environments.

Critic parameters are updated based on robust tracking variables:
\begin{align}
\varphi_i^{(k+1)} = \varphi_i^{(k)} - \alpha_{cag} \cdot \mathbf{y}_i^{(k+1)},
\end{align}
where $\alpha_{cag}$ is the aggregation learning rate for the Critic network.

Algorithm~\ref{alg:robust_decentralized_agg} presents the complete dual-track Non-IID robust decentralized aggregation. It first constructs the communication topology in the initialization phase (Lines 2-7) by selecting $d_{max}$ lowest-latency neighbors through RTT measurements for each silo, and initializes gradient fingerprint buffers and tracking variables. In the local training phase (Lines 10-18), each silo performs $\Omega$ rounds of Actor-Critic training in parallel (invoking Algorithm~\ref{alg:silo_local_training}), while collecting resource feature gradients and maintaining a sliding window of size $K$, then computes the gradient fingerprint as a compact representation of policy preferences upon training completion. In the aggregation phase (Lines 20-42), each silo first exchanges gradient fingerprints, model parameters, and tracking variables with neighbors (Lines 22-24), then computes similarity based on gradient fingerprints and performs robust anomaly detection through MAD to identify suspicious neighbors (Lines 26-32). For the Actor network, similarity-weighted aggregation is employed, where aggregation weights are assigned based on policy similarity after excluding anomalous neighbors (Lines 34-35); for the Critic network, the gradient tracking mechanism is adopted, achieving optimization momentum sharing through robust gradient averaging and tracking variable updates (Lines 37-38). Finally, the communication topology is dynamically optimized based on neighbor recommendations and RTT improvements (Line 41).

\begin{algorithm}[htbp]
\footnotesize
\caption{Dual-Track Non-IID Robust Decentralized Aggregation}
\label{alg:robust_decentralized_agg}
\begin{algorithmic}[1]
\REQUIRE Silos $\mathcal{Z} = \{z_1, z_2, \ldots, z_M\}$, aggregation period $\Omega$, neighbor size $d_{max}$, similarity window $K$, temperature $\nu$, robustness parameter $\xi$, aggregation rates $\alpha_{\mathrm{agg}}, \alpha_{\mathrm{cag}}$
\ENSURE Updated Actor parameters $\{\phi_i\}$ and Critic parameters $\{\varphi_i\}$

\STATE // Initialization: Network Topology Construction
\FOR{each silo $z_i \in \mathcal{Z}$}
    \STATE Sample $k_{\text{sample}}$ candidates, measure RTT
    \STATE Select $d_{\max}$ lowest-latency neighbors $\rightarrow \mathcal{H}_i(0)$
    \STATE Initialize gradient fingerprint buffer $\mathcal{B}_i^{\text{grad}} = \{\}$
    \STATE Initialize tracking variable $\mathbf{y}_i^{(0)} = \mathbf{0}$
\ENDFOR

\FOR{aggregation round $r = 1, 2, \ldots$}
    \STATE // Local Training Phase ($\Omega$ rounds)
    \FOR{each silo $z_i$ \textbf{in parallel}}
        \FOR{local step $k = 1, 2, \ldots, \Omega$}
            \STATE Perform local Actor-Critic training (Algorithm~\ref{alg:silo_local_training})
            \STATE Collect resource feature gradient: $\nabla_{\psi_{\mathrm{resource}}(n_k)} u_\phi(s_k,n_k)$
            \STATE Add to buffer: $\mathcal{B}_i^{\text{grad}} \leftarrow \mathcal{B}_i^{\text{grad}} \cup \{\nabla_{\psi_{\mathrm{resource}}(n_k)} u_\phi(s_k,n_k)\}$
            \STATE Keep only most recent $K$ gradients in $\mathcal{B}_i^{\text{grad}}$
        \ENDFOR
        \STATE Compute gradient fingerprint: $\mathbf{g}_i = \frac{1}{K}\sum_{\mathbf{g} \in \mathcal{B}_i^{\text{grad}}} \mathbf{g}$
    \ENDFOR
    
    \STATE // Aggregation Phase
    \FOR{each silo $z_i$}
        \STATE // Information Exchange with All Neighbors
        \FOR{each neighbor $z_j \in \mathcal{H}_i$}
            \STATE Exchange gradient fingerprints $\mathbf{g}_j$, Actor parameters $\phi_j^{(r)}$, Critic gradients $\mathbf{d}_j^{(r)}$, tracking variables $\mathbf{y}_j^{(r)}$, and neighbor recommendations
        \ENDFOR
        
        \STATE // Similarity Assessment and Anomaly Detection
        \FOR{each neighbor $z_j \in \mathcal{H}_i$}
            \STATE Compute similarity: $\mathrm{sim}_{ij} = \frac{\mathbf{g}_i^\top \mathbf{g}_j}{\|\mathbf{g}_i\| \|\mathbf{g}_j\|}$
        \ENDFOR
        \STATE Compute median similarity: $\text{med}_i = \text{median}(\{\mathrm{sim}_{ij} : j \in \mathcal{H}_i\})$
        \STATE Compute MAD: $\text{MAD}_i = \text{median}(\{|\mathrm{sim}_{ij} - \text{med}_i| : j \in \mathcal{H}_i\})$
        \STATE Set anomaly threshold: $\tau_{\mathrm{sim}} = \text{med}_i - \xi \cdot \text{MAD}_i$
        \STATE Identify anomalous neighbors: $\mathcal{O}_i = \{j \in \mathcal{H}_i : \mathrm{sim}_{ij} < \tau_{\mathrm{sim}}\}$
        
        \STATE // Actor Network Aggregation
        \STATE Compute aggregation weights: $w_{ij} = \frac{\exp(\mathrm{sim}_{ij}/\nu)}{\sum_{j' \in \mathcal{H}_i \setminus \mathcal{O}_i} \exp(\mathrm{sim}_{ij'}/\nu)}$
        \STATE Update Actor parameters: $\phi_i^{(r+1)} = (1-\alpha_{\mathrm{agg}}) \cdot \phi_i^{(r)} + \alpha_{\mathrm{agg}} \cdot \sum_{j \in \mathcal{H}_i \setminus \mathcal{O}_i} w_{ij} \cdot \phi_j^{(r)}$
        
        \STATE // Critic Network Aggregation
        \STATE Compute robust gradients: $\tilde{\mathbf{d}}_i^{(r)} = \frac{1}{|\mathcal{H}_i \setminus \mathcal{O}_i| + 1} \sum_{j \in (\mathcal{H}_i \setminus \mathcal{O}_i) \cup \{i\}} \mathbf{d}_j^{(r)}$
        \STATE Update tracking variables: $\mathbf{y}_i^{(r+1)} = \sum_{j \in \mathcal{H}_i \cup \{i\}} c_{ij} (\mathbf{y}_j^{(r)} + \tilde{\mathbf{d}}_j^{(r+1)} - \tilde{\mathbf{d}}_j^{(r)})$
        \STATE Update Critic parameters: $\varphi_i^{(r+1)} = \varphi_i^{(r)} - \alpha_{\mathrm{cag}} \cdot \mathbf{y}_i^{(r+1)}$
        
        \STATE // Dynamic Neighbor Optimization
        \STATE Optimize neighbor set using received recommendations and RTT improvements
    \ENDFOR
\ENDFOR
\end{algorithmic}
\end{algorithm}

\section{Performance Evaluation}
\label{evaluation}
This section presents a comprehensive experimental evaluation of DeFRiS in realistic silo-cooperative IoT environments. We first describe the experimental setup and hyperparameter configuration, then evaluate DeFRiS across five dimensions: convergence performance, ablation study, QoS guarantee, scalability, and robustness in adversarial environments.

\subsection{Experiment Setup}
This subsection describes the distributed silo-cooperative testbed, IoT application workloads, and baseline approaches.

\subsubsection{Practical Experiment Environment}
To validate the effectiveness of DeFRiS in realistic silo-cooperative IoT environments, we establish a distributed experimental testbed comprising 20 autonomous computing silos, each with different resource configurations and workload characteristics. Each silo operates as an independent management entity with complete autonomy over its local resource scheduling decisions while participating in the decentralized federated learning protocol for cooperative optimization.

Each computing silo contains a heterogeneous mix of IoT devices, edge servers, and cloud instances, forming a realistic three-tier computing hierarchy:

\textbf{Cloud Layer}: We deploy cloud instances from diverse providers to ensure environmental heterogeneity, including Nectar cloud instances (AMD EPYC processors, configurations ranging from 2 cores @2.0GHz 8GB RAM to 32 cores @2.0GHz 128GB RAM), AWS cloud instances (Intel Xeon processors, configurations ranging from 1 core @2.4GHz 1GB RAM to 16 cores @2.5GHz 64GB RAM), and Microsoft Azure cloud instances (Intel Xeon processors, configurations ranging from 1 core @2.3GHz 1GB RAM to 24 cores @2.4GHz 96GB RAM).

\textbf{Edge Layer}: We configure various edge computing devices, including M1 Pro processor-based devices (8 cores, 16GB RAM), Intel Core i7 processor-equipped devices (8 cores @2.3GHz, 16GB RAM), Intel Core i9 processor devices (8 cores @2.5GHz, 32GB RAM), and Intel Core i5 processor devices (6 cores @2.8GHz, 8GB RAM) with different configurations.

\textbf{IoT Device Layer}: We deploy Raspberry Pi devices (Pi OS, Broadcom BCM2837 quad-core @1.2GHz, 1GB RAM), virtual machines, and Docker containers equipped with cameras and IP cameras. Some IoT devices are equipped with limited battery capacity (simulated as 10,000-50,000 Joules), while others are connected to stable power supplies.

To reflect inter-silo heterogeneity, we configure different resource combinations for different silos: Silos 1-5 primarily contain cloud computing resources (70\% cloud instances), Silos 6-10 are edge-computing dominant (60\% edge nodes), Silos 11-15 adopt balanced configurations (evenly distributed resources across all tiers), and Silos 16-20 are IoT-device dense (over 50\% IoT devices).

Network connections exhibit realistic latency and bandwidth variations, reflecting real-world deployment scenarios. The latency between IoT devices and edge devices ranges from 1-6ms with a bandwidth of 10-25 MB/s. Network characteristics between IoT devices and cloud instances vary by cloud service provider, with latency ranging from 6-25ms and bandwidth from 14-22MB/s. Communication latency between edge devices and cloud instances ranges from 6-25ms with a bandwidth of 15-22MB/s.

For silo-cooperative learning communication, each silo maintains up to 5 neighbor connections ($d_{max} = 5$), dynamically selecting optimal neighbors based on RTT measurements. For energy monitoring, we use the eco2AI \cite{budennyy2022eco2ai} toolkit to implement accurate real-time power measurement. In the response time objective function Eq. \ref{eq:time}, CVaR risk control parameters are set to $\alpha = 0.95$ and $\beta = 0.3$, ensuring tail risk control while optimizing average performance. In the joint optimization objective function Eq. \ref{eq:target}, we set weight parameters $\lambda_{RT}^{(i)} = 0.5$ and $\lambda_{Energy}^{(i)} = 0.5$ to balance response time and energy consumption optimization objectives. 

\subsubsection{IoT Application Workloads}
To comprehensively evaluate DeFRiS across realistic silo-cooperative scenarios, we deploy a diverse suite of IoT applications adapted from production deployment requirements:

\begin{itemize}
  \item \textit{VoiceHomeController}: speech-to-text and intent recognition for smart home automation, combining \texttt{torchaudio} \cite{yang2022torchaudio} and a lightweight Transformer decoder; workload scaled by audio-chunk length.
  
  \item \textit{IndustrialNoiseMonitor}: acoustic amplitude tracking for factory equipment health monitoring, implemented with \texttt{librosa} \cite{mcfee2015librosa}; workload scaled by analysis-window length.
  
  \item \textit{DriverFatigueMonitor}: cascaded face-and-eye detection with fatigue state classification for automotive safety systems, using \texttt{OpenCV} \cite{bradski2000opencv} and \texttt{dlib} \cite{king2009dlib}; workload scaled by input resolution and cascade depth.
  
  \item \textit{MedicalImagePreprocessor}: batch medical image filtering, enhancement, and standardization for telehealth diagnostics, utilizing \texttt{PIL} \cite{clark2015pillow} and \texttt{OpenCV} \cite{bradski2000opencv}; workload scaled by batch size and processing pipeline depth.
  
  \item \textit{GestureInteractionService}: HSV-based hand-gesture tracking for AR/VR interaction systems using \texttt{OpenCV} \cite{bradski2000opencv}; workload scaled by frame rate and tracking-window size.
  
  \item \textit{CustomerFeedbackAnalyzer}: real-time sentiment analysis on customer reviews for retail analytics, using \texttt{TextBlob} \cite{loria2018textblob} and \texttt{NLTK} \cite{bird2006nltk}; workload scaled by text-block size.
  
  \item \textit{ElderlyFallDetection}: pose-estimation-based fall detection for elderly care monitoring, implemented with \texttt{TensorFlow~Lite} \cite{abadi2016tensorflow, tflite2017}; workload scaled by sampling rate and model capacity.
  
  \item \textit{TrafficSignRecognition}: video-based traffic sign detection and text recognition for ADAS systems, combining \texttt{EasyOCR} \cite{easyocr2020} and \texttt{OpenCV} \cite{bradski2000opencv}; workload scaled by clip length and model precision.
  
  \item \textit{EdgeDataTransmitter}: adaptive data compression for bandwidth-constrained IoT gateways, using \texttt{zlib} \cite{gailly1995zlib}/\texttt{gzip} \cite{rfc1952}; workload scaled by file size and compression level.
  
  \item \textit{SecuritySurveillance}: real-time face detection for access control and intrusion monitoring, implemented with \texttt{OpenCV} \cite{bradski2000opencv} and \texttt{MediaPipe} \cite{lugaresi2019mediapipe}; workload scaled by frame resolution and detection frequency.
\end{itemize}

To reflect workload heterogeneity in real-world deployments, different silo types tend to receive applications matching their resource profiles: cloud-dominant silos (1-5) more frequently receive compute-intensive tasks (e.g., MedicalImagePreprocessor processing high-resolution imagery, CustomerFeedbackAnalyzer analyzing large-scale text datasets); edge-dominant silos (6-10) more often execute latency-sensitive applications (e.g., DriverFatigueMonitor processing real-time video streams, VoiceHomeController providing instant voice responses); balanced silos (11-15) receive a mixture of application types; IoT-dense silos (16-20) primarily deploy energy-optimized variants (e.g., ElderlyFallDetection using pruned lightweight pose models, IndustrialNoiseMonitor employing downsampled audio processing to conserve energy). By systematically varying workload parameters (e.g., video resolution from 480p to 4K, audio sampling rate from 8kHz to 48kHz, batch size from 10 to 500, model complexity from tiny to large), we generate a comprehensive set of application instances that stress-test DeFRiS across diverse IoT scheduling scenarios.

\subsubsection{Baseline Approaches}
We implement DeFRiS using ReinFog\cite{wang2025reinfog}, a modular framework for DRL-based resource management that supports both centralized and distributed learning techniques development and integration in practical edge-fog-cloud environments. To comprehensively evaluate DeFRiS, we compare it against four representative state-of-the-art approaches:
\begin{itemize}
  \item \textbf{MCM-FDRL} \cite{10949717}: A federated DRL framework where independent clients locally train DQN models and update a global model via periodic parameter averaging. We adapt it to multi-silo scenarios by treating each silo as a client and modifying the action design and reward functions.
  
  \item \textbf{TF-DDRL} \cite{wang2025tf}: A distributed DRL approach employing Transformer and IMPALA framework for asynchronous training. We extend its architecture to multi-silo scenarios by deploying distributed actors across heterogeneous silos to asynchronously feed the central learner.
  
  \item \textbf{VCPN} \cite{11039641}: A multi-agent DRL approach based on MADDPG using centralized training with shared replay buffers and decentralized execution. We adapt it to multi-silo scenarios by treating each silo as an agent and modifying the state representation and reward functions.
  
  \item \textbf{D3QN-SEIDEL} \cite{11164488}: A single-agent DRL approach based on D3QN optimizing task offloading through separate value and advantage streams. We deploy independent agents per silo and adapt the state-action space and reward design for each local environment.
\end{itemize}

These baselines are strategically selected to cover the full spectrum of coordination paradigms in IoT scheduling: MCM-FDRL represents federated learning with global parameter aggregation \cite{10618900,10949717}; TF-DDRL exemplifies parallelized centralized learning where distributed actors asynchronously collect experience for a shared policy \cite{10540320,zhang2024lsia3cs,wang2025tf}; VCPN adopts the multi-agent paradigm with Centralized Training and Decentralized Execution (CTDE) to handle inter-agent interactions \cite{10843979,10960753,11039641}; and D3QN-SEIDEL serves as the baseline for independent single-agent learning without external coordination \cite{10380323,10989563,10848209,wang2024deep,11164488}. This diverse selection enables a comprehensive evaluation of DeFRiS across distinct methodological approaches.

\subsection{Hyperparameter Configuration}
To ensure optimal performance, we conduct systematic hyperparameter tuning for DeFRiS and all baseline approaches using grid search. All approaches share identical network architectures where applicable to ensure fair comparison. Table~\ref{tab:hyperparameters} summarizes the key hyperparameters for DeFRiS.
\begin{table}[htbp]
\centering
\caption{Hyperparameter Configuration}
\label{tab:hyperparameters}
\renewcommand{\arraystretch}{0.9}
\footnotesize
\begin{tabular}{lc}
\toprule
\textbf{Parameter} & \textbf{Value} \\
\midrule
\multicolumn{2}{l}{\textit{Objective Function}} \\
CVaR risk level ($\alpha$) & 0.95 \\
CVaR weight ($\beta$) & 0.3 \\
Response time weight ($\lambda_{RT}^{(i)}$) & 0.5 \\
Energy consumption weight ($\lambda_{Energy}^{(i)}$) & 0.5 \\
\midrule
\multicolumn{2}{l}{\textit{Local Training}} \\
Discount factor ($\gamma$) & 0.99 \\
Actor learning rate ($\alpha_{\phi}$) & 3e-4 \\
Critic learning rate ($\alpha_{\varphi}$) & 1e-3 \\
GAE parameter ($\lambda$) & 0.95 \\
Clipping parameter ($\epsilon$) & 0.2 \\
\midrule
\multicolumn{2}{l}{\textit{Network Architecture}} \\
Hidden layers & 2 \\
Hidden units & 128-512 \\
Activation function & ReLU \\
\midrule
\multicolumn{2}{l}{\textit{Federated Aggregation}} \\
Maximum neighbors ($d_{max}$) & 5 \\
Candidate sampling size ($k_{sample}$) & 10 \\
Gradient fingerprint window ($K$) & 100 \\
Similarity temperature ($\nu$) & 0.1 \\
Robustness parameter ($\xi$) & 3.0 \\
Actor aggregation rate ($\alpha_{\mathrm{agg}}$) & 0.3 \\
Critic aggregation rate ($\alpha_{\mathrm{cag}}$) & 0.1 \\
\bottomrule
\end{tabular}
\end{table}

\subsection{Experimental Results and Analysis}
We evaluate DeFRiS through convergence performance, ablation study, QoS guarantee analysis, scalability evaluation, and robustness in adversarial environments. All experiments are repeated 10 times with different random seeds, and the reported results represent the mean performance averaged over all runs.

\subsubsection{Convergence Performance}
\begin{figure*}[t]
    \centering
    \begin{subfigure}[b]{0.32\textwidth}
        \centering
        \includegraphics[width=\textwidth]{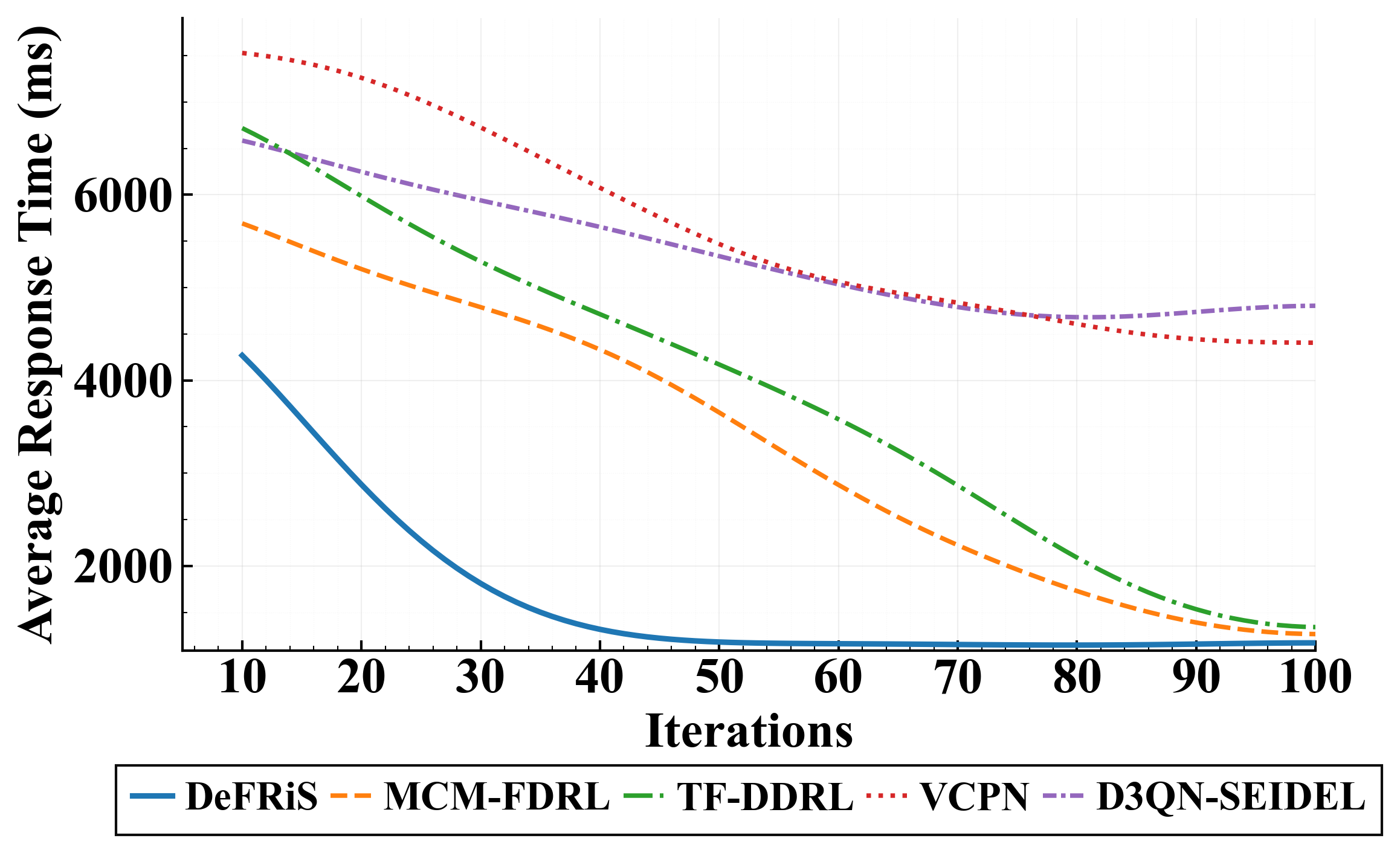}
        \caption{Response Time}
        \label{fig:conv_time}
    \end{subfigure}
    \hfill
    \begin{subfigure}[b]{0.32\textwidth}
        \centering
        \includegraphics[width=\textwidth]{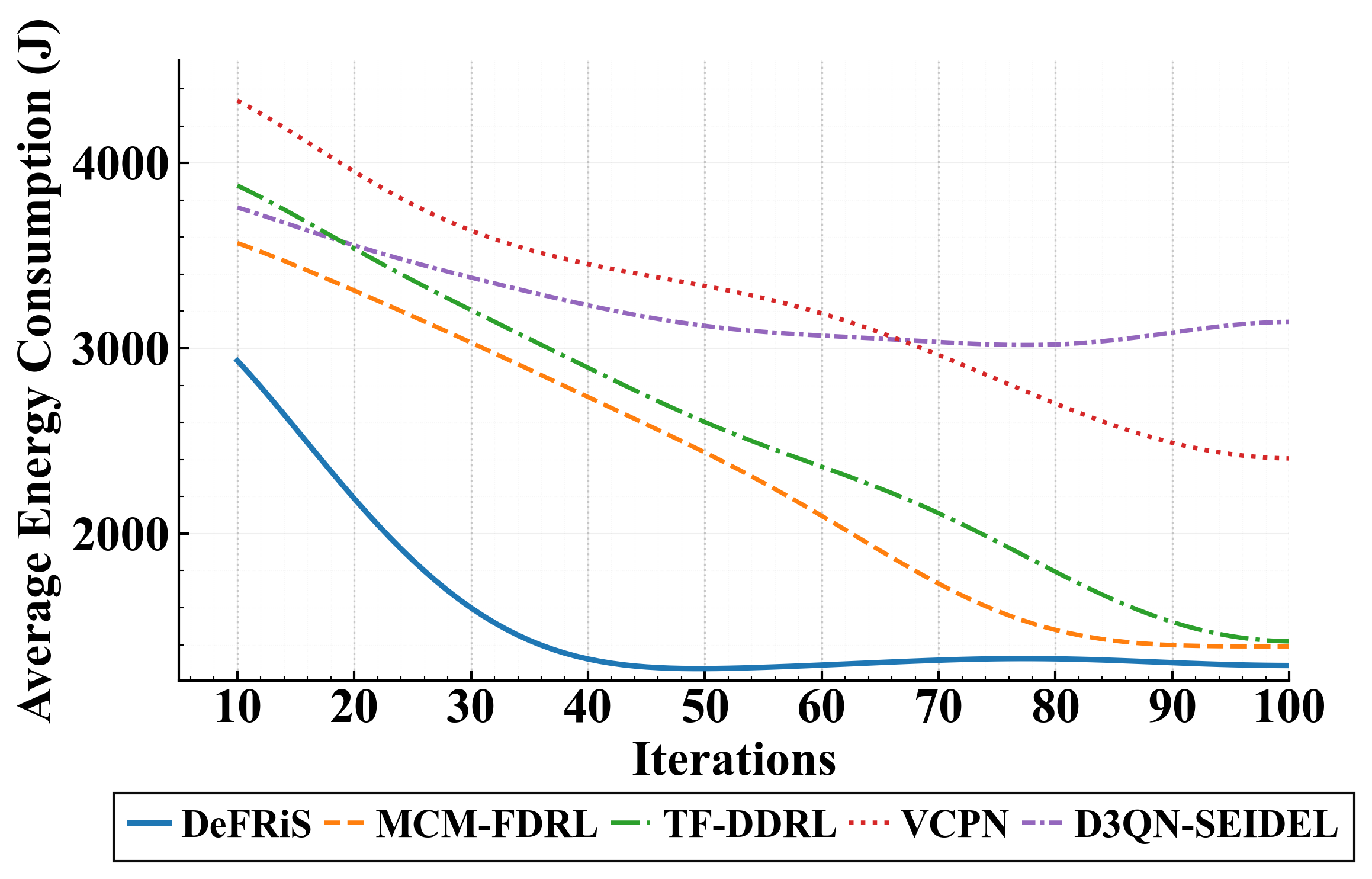}
        \caption{Energy Consumption}
        \label{fig:conv_energy}
    \end{subfigure}
    \hfill
    \begin{subfigure}[b]{0.32\textwidth}
        \centering
        \includegraphics[width=\textwidth]{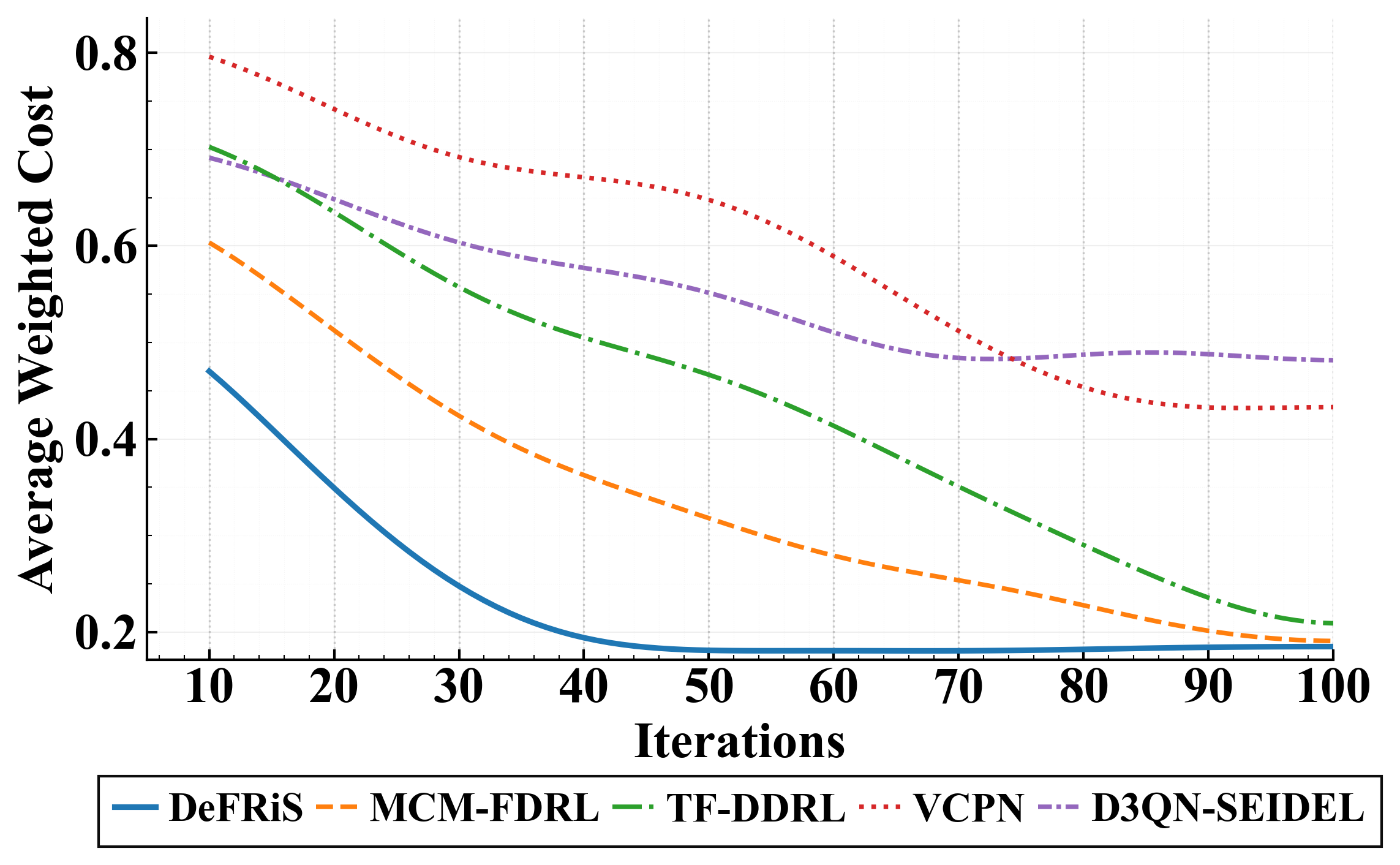}
        \caption{Weighted Cost}
        \label{fig:conv_cost}
    \end{subfigure}
    \caption{Convergence performance comparison across 100 training iterations.}
    \label{fig:convergence}
\end{figure*}

Figure~\ref{fig:convergence} presents the convergence behavior of all approaches across 100 training iterations, evaluated in terms of response time, energy consumption, and weighted cost. All metrics represent the average performance across 20 silos, reflecting system-level overall performance. DeFRiS consistently demonstrates superior convergence speed and final performance across all three metrics.

\textbf{Response time optimization.} As shown in Figure~\ref{fig:conv_time}, DeFRiS achieves a final average response time of around 1170 ms, representing 6.4\% and 13.3\% improvements over MCM-FDRL (around 1250 ms) and TF-DDRL (around 1350 ms), respectively, and substantial gains of 73.4\% and 75.6\% compared to VCPN (around 4400 ms) and D3QN-SEIDEL (around 4800 ms). DeFRiS exhibits a rapid descent during the first 40 iterations and stabilizes near the optimal solution by iteration 50. In contrast, MCM-FDRL and TF-DDRL require around 100 iterations to achieve comparable stability, while VCPN and D3QN-SEIDEL remain trapped at substantially higher response times throughout training, reflecting optimization difficulties in multi-silo Non-IID environments.

\textbf{Energy efficiency.} Figure~\ref{fig:conv_energy} demonstrates that DeFRiS achieves a final energy consumption of around 1290 J, saving 7.2\% and 9.2\% compared to MCM-FDRL (around 1390 J) and TF-DDRL (around 1420 J), respectively, and achieving substantial reductions of 46.3\% and 58.4\% over VCPN (around 2400 J) and D3QN-SEIDEL (around 3100 J). DeFRiS exhibits rapid energy optimization during the first 40 iterations and stabilizes by iteration 50. In contrast, MCM-FDRL and TF-DDRL require nearly 100 iterations to converge, while VCPN cannot converge within 100 iterations and maintains consistently high energy consumption throughout training. Notably, D3QN-SEIDEL exhibits an energy rebound phenomenon after iteration 80, demonstrating the instability of independent learning without cross-silo knowledge transfer.

\textbf{Weighted cost performance.} The joint optimization metric in Figure~\ref{fig:conv_cost} shows that DeFRiS achieves the lowest final cost of around 0.185, outperforming MCM-FDRL (around 0.191), TF-DDRL (around 0.209), VCPN (around 0.433), and D3QN-SEIDEL (around 0.482) by 3.0\%, 11.5\%, 57.3\%, and 61.6\%, respectively. Similarly, DeFRiS demonstrates rapid optimization during the first 40 iterations and stabilizes by iteration 50. In contrast, MCM-FDRL and TF-DDRL require the full 100 iterations to approach convergence, while VCPN and D3QN-SEIDEL stabilize at substantially higher cost levels, highlighting the critical importance of effective knowledge sharing and action space design in heterogeneous multi-silo environments.

DeFRiS's superior performance stems from the synergistic effect of three core components. First, the action-space-agnostic policy enables parameter sharing across heterogeneous silos through candidate scoring mechanisms, allowing silos with different resource configurations to effectively transfer scheduling knowledge. Second, silo-optimized local learning combines GAE with clipped policy updates, addressing both the credit assignment problem under sparse rewards and ensuring training stability, thereby achieving rapid and smooth convergence. Finally, Non-IID robust decentralized aggregation employs a dual-track aggregation strategy that selectively transfers policy knowledge through gradient fingerprints while preserving local accuracy of value functions through gradient tracking. 

\subsubsection{Ablation Study}
\begin{figure*}[t]
    \centering
    \begin{minipage}[b]{0.32\textwidth}
        \centering
        \includegraphics[width=\textwidth]{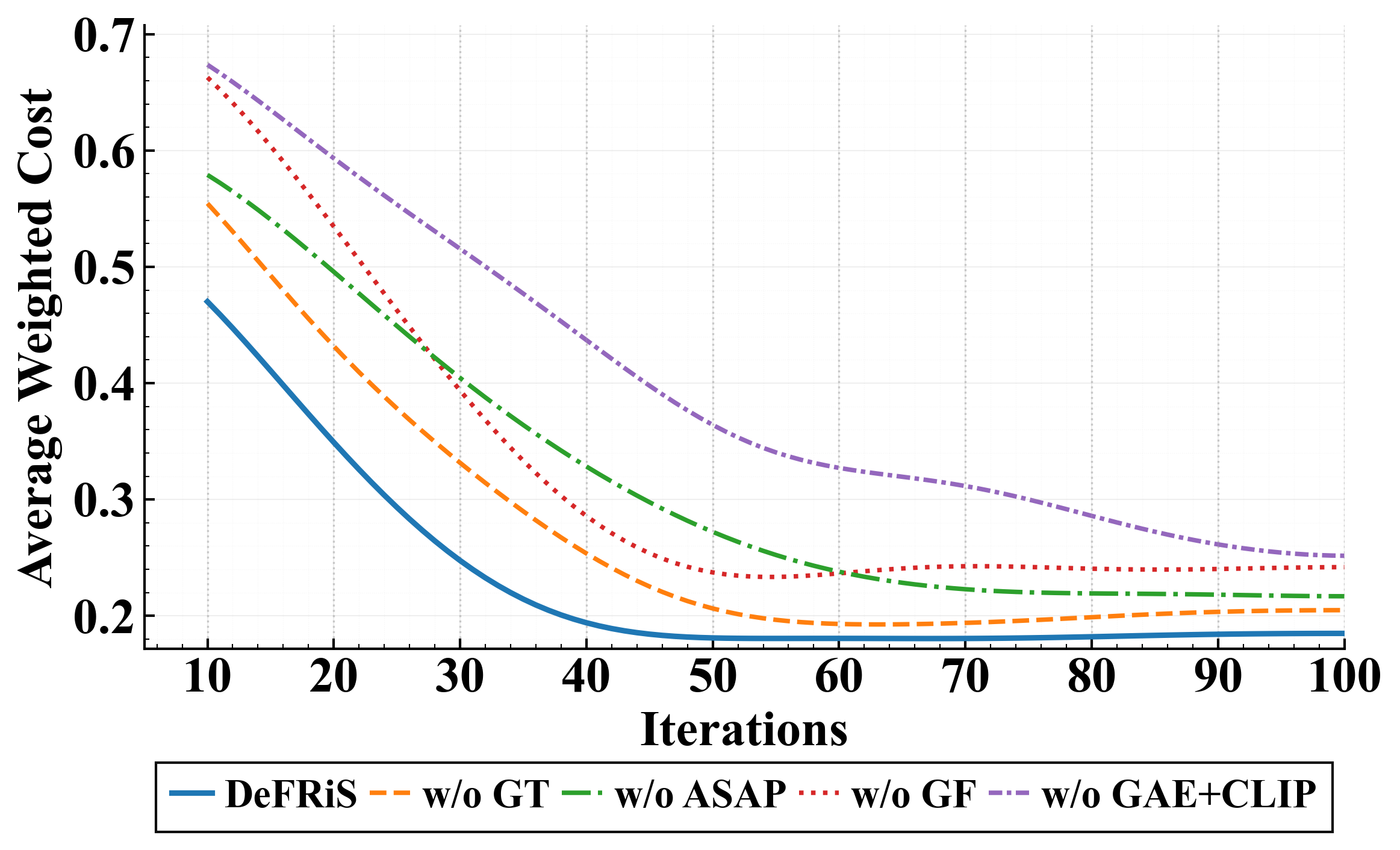}
        \captionof{figure}{Ablation study results showing performance degradation when removing individual components of DeFRiS.}
        \label{fig:ablation}
    \end{minipage}
    \hfill
    \begin{minipage}[b]{0.66\textwidth}
        \centering
        \begin{subfigure}[b]{0.48\textwidth}
            \centering
            \includegraphics[width=\textwidth]{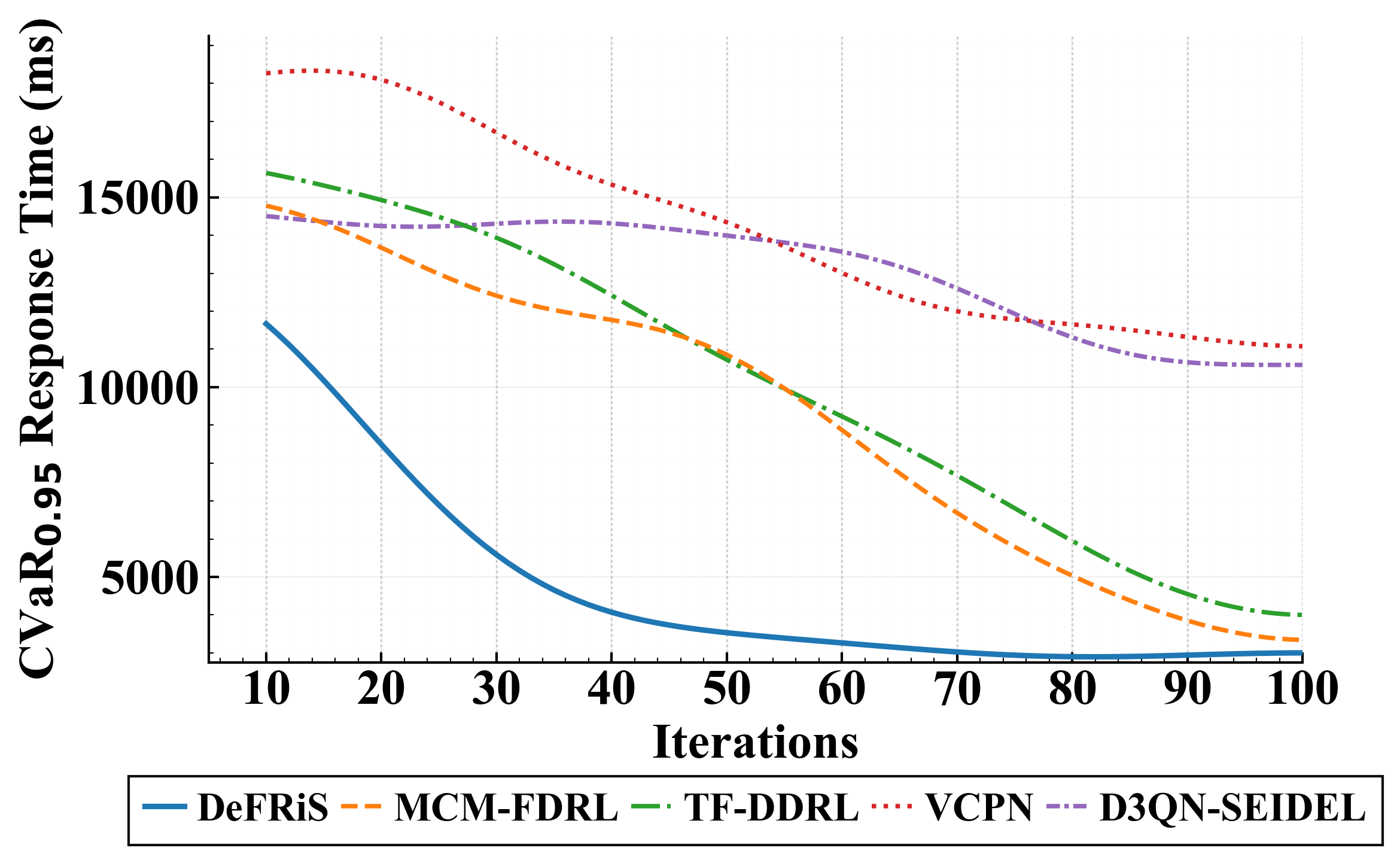}
            \caption{CVaR$_{0.95}$ Response Time}
            \label{fig:qos_cvar}
        \end{subfigure}
        \hfill
        \begin{subfigure}[b]{0.48\textwidth}
            \centering
            \includegraphics[width=\textwidth]{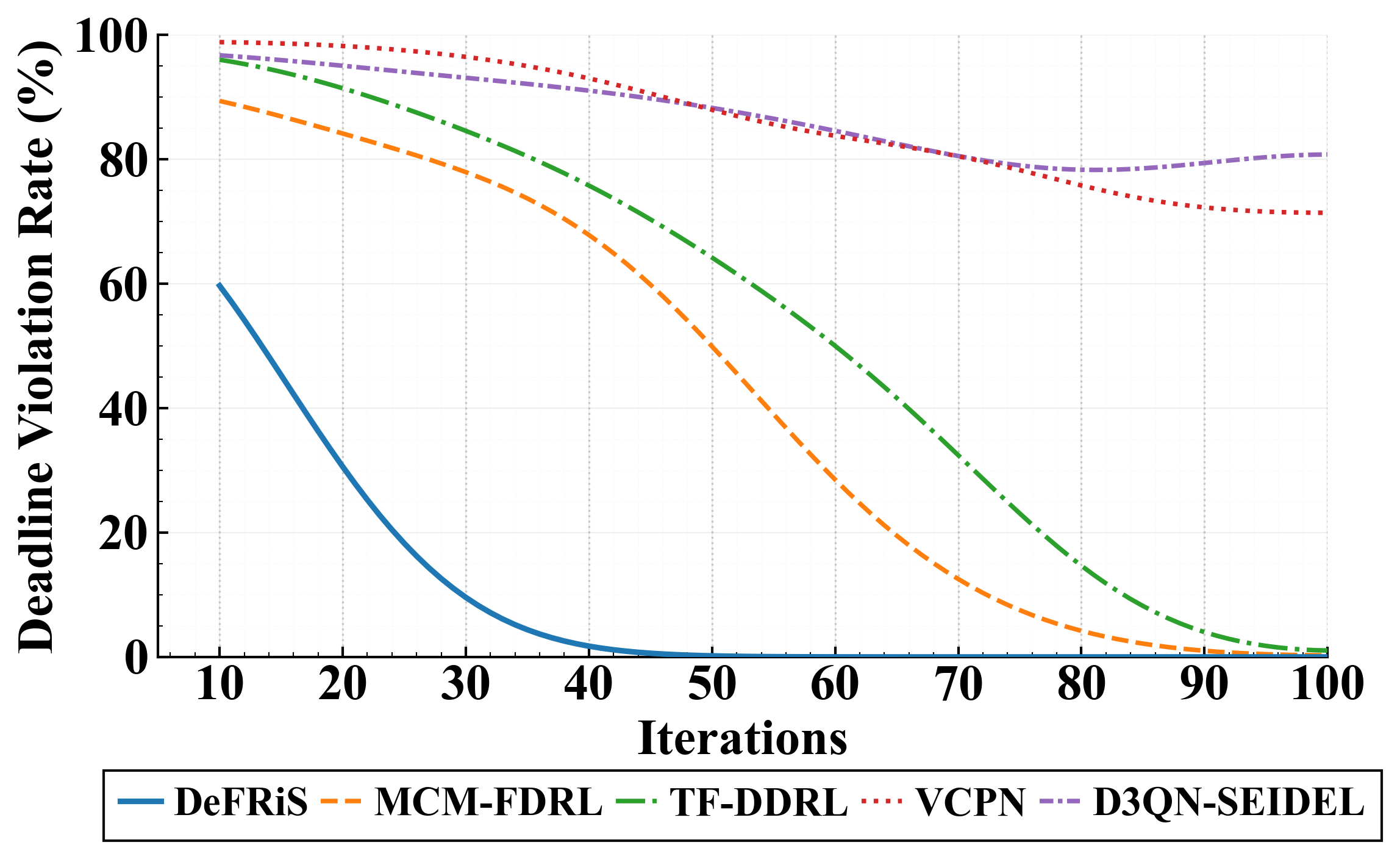}
            \caption{Deadline Violation Rate}
            \label{fig:qos_dvr}
        \end{subfigure}
        \captionof{figure}{QoS guarantee performance comparison.}
        \label{fig:qos}
    \end{minipage}
\end{figure*}

To validate the individual contribution of each core component in DeFRiS, we conduct systematic ablation experiments. Specifically, we construct four variants that respectively remove the action-space-agnostic policy (w/o ASAP adopts fixed-dimension output layers with masking to handle heterogeneous action spaces), GAE+CLIP mechanism (w/o GAE+CLIP adopts vanilla policy gradient), gradient fingerprint-based Actor aggregation (w/o GF adopts simple parameter averaging instead), and gradient tracking-based Critic aggregation (w/o GT adopts simple parameter averaging instead). As shown in Figure~\ref{fig:ablation}, the full DeFRiS achieves a final cost of 0.185 and significantly outperforms all variants at every stage of training.

Removing the GAE+CLIP mechanism causes the most severe performance loss, with the final cost degrading to 0.252 (36.2\% degradation). This variant also exhibits the slowest convergence throughout training, reaching 0.674 at iteration 10 and remaining at 0.516 at iteration 30 while full DeFRiS has descended to 0.248. This validates that both GAE's credit assignment and clipped updates' stability guarantees are indispensable in sparse reward environments, as their absence severely impairs learning efficiency and produces low-quality parameters that contaminate federated aggregation.

Removing gradient fingerprint leads to the second largest performance loss at 0.242 (30.8\% degradation). Starting at 0.663 at iteration 10, this variant remains at 0.394 at iteration 30 and plateaus around 0.24 after iteration 50, reflecting the harm of blind aggregation in Non-IID environments. Simple parameter averaging forcibly fuses incompatible policy knowledge from heterogeneous silos, while gradient fingerprint's similarity-aware selective aggregation enables effective knowledge transfer matching each silo's environment.

Removing action-space-agnostic policy results in 17.3\% degradation to 0.217. At iteration 10, this variant reaches 0.579, and at iteration 30 remains at 0.405 versus the full version's 0.248, demonstrating significantly slower convergence. ASAP's candidate scoring mechanism enables more efficient representation learning through its unified resource feature space, achieving faster policy optimization and more effective knowledge aggregation across heterogeneous silos throughout training.

Removing gradient tracking degrades performance to 0.205 (10.8\% degradation). This variant reaches 0.555 at iteration 10 and 0.322 at iteration 30, consistently falling behind full DeFRiS throughout training. This confirms that simple Critic parameter averaging destroys accurate modeling of local state distributions, while gradient tracking successfully balances cooperative learning and local adaptation by sharing optimization momentum while maintaining personalized value functions.

\subsubsection{QoS Guarantee Analysis}
To evaluate the performance of each approach in service quality assurance, we analyze the convergence behavior of two critical metrics. Figure~\ref{fig:qos_cvar} shows CVaR$_{0.95}$ response time, measuring tail latency risk at 95\% confidence level, while Figure~\ref{fig:qos_dvr} shows deadline violation rate, measuring the proportion of tasks failing to meet latency requirements. DeFRiS achieves fast convergence and optimal performance on both metrics, validating its effectiveness in guaranteeing QoS requirements for mission-critical tasks.

\textbf{Tail latency optimization.} As shown in Figure~\ref{fig:qos_cvar}, DeFRiS's CVaR$_{0.95}$ response time rapidly decreases from around 11700 ms at iteration 10 to around 5800 ms at iteration 30 (50.4\% reduction), ultimately converging to around 3000 ms. In contrast, while MCM-FDRL and TF-DDRL also achieve reductions, their convergence speeds are significantly slower than DeFRiS, finally stabilizing at around 3350 ms and 4000 ms, respectively. More notably, VCPN and D3QN-SEIDEL maintain extremely high tail latencies throughout training, ultimately reaching around 11000 ms. DeFRiS reduces tail latency by 10.4\% compared to the best baseline MCM-FDRL, an improvement crucial for latency-sensitive critical tasks, as tail latency often determines worst-case performance guarantees.

\textbf{Deadline violation rate optimization.} Figure~\ref{fig:qos_dvr} demonstrates even more significant performance differences. DeFRiS achieves a near 0\% violation rate by iteration 50. MCM-FDRL and TF-DDRL exhibit slower improvement rates, requiring over 90 iterations to decrease to the same level. VCPN and D3QN-SEIDEL perform particularly poorly, with violation rates approaching 100\% at iteration 10, and despite limited improvement throughout training, ultimately maintaining high levels of around 70\% and 80\%. 

These results validate DeFRiS's ability to satisfy strict QoS constraints in production environments. By achieving near 0\% deadline violations and significantly lower tail latencies, DeFRiS demonstrates the critical importance of effective cross-silo coordination in meeting real-time service requirements that baseline approaches consistently fail to guarantee.

\subsubsection{Scalability Evaluation}
\begin{figure}[htbp]
    \centering
    \includegraphics[width=0.48\textwidth]{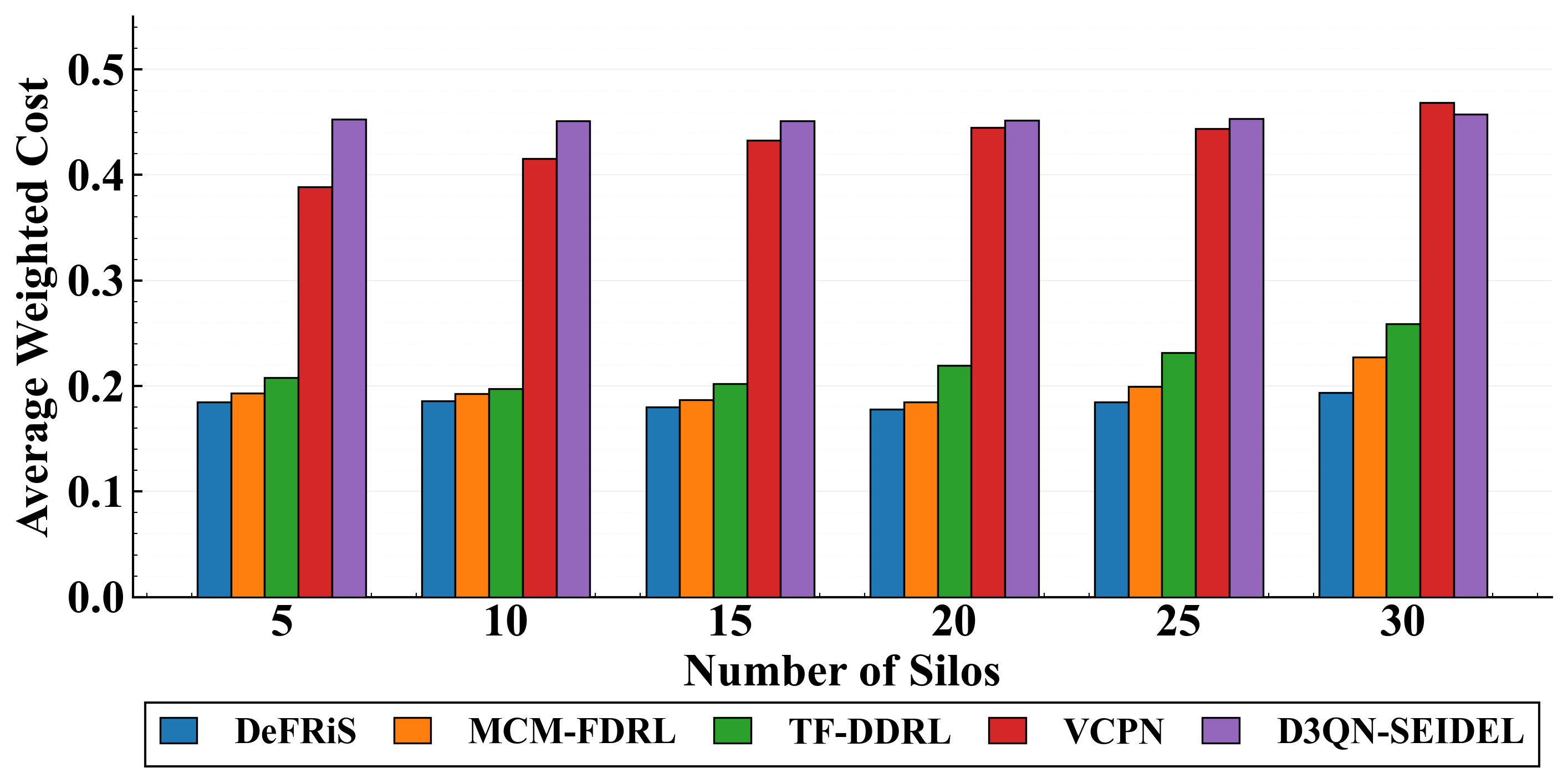}
    \caption{Scalability evaluation showing average weighted cost across different numbers of silos.}
    \label{fig:scalability}
\end{figure}

To evaluate the scalability of DeFRiS in large-scale distributed deployments, we systematically vary the number of independent silos from 5 to 30, testing the final performance of each approach at different scales. Figure~\ref{fig:scalability} presents the scalability performance of all approaches.

Within the range of 5 to 20 silos, DeFRiS maintains a stable average weighted cost (approximately 0.182), 3.7\% and 12.1\% lower than MCM-FDRL (approximately 0.189) and TF-DDRL (approximately 0.207), respectively, and 56.7\% and 59.7\% lower than VCPN (approximately 0.420) and D3QN-SEIDEL (approximately 0.452), respectively. When the scale expands to 30 silos, performance divergence becomes significant: DeFRiS increases moderately to 0.194 (a 6.6\% increase), while MCM-FDRL, TF-DDRL, and VCPN surge dramatically to 0.227 (20.1\% increase), 0.259 (25.1\% increase), and 0.468 (11.4\% increase), respectively, and D3QN-SEIDEL remains at 0.458 (1.3\% increase). This demonstrates that DeFRiS achieves over 3 times better performance retention compared to the best-performing baseline (MCM-FDRL) as the system scales. DeFRiS's superior scalability stems from its decentralized architecture that avoids centralized bottlenecks, gradient fingerprint and gradient tracking-based aggregation that enables robust policy transfer while preserving local adaptation, and dynamic neighbor optimization that controls communication overhead. In contrast, the centralized architectures of MCM-FDRL, TF-DDRL, and VCPN all face centralized coordination bottlenecks and Non-IID noise accumulation as scale increases, leading to sharp performance degradation. While D3QN-SEIDEL's independent learning avoids scale overhead, it cannot benefit from cooperation, maintaining a high cost level.

\subsubsection{Robustness in Adversarial Environments}
\begin{figure}[htbp]
    \centering
    \includegraphics[width=0.48\textwidth]{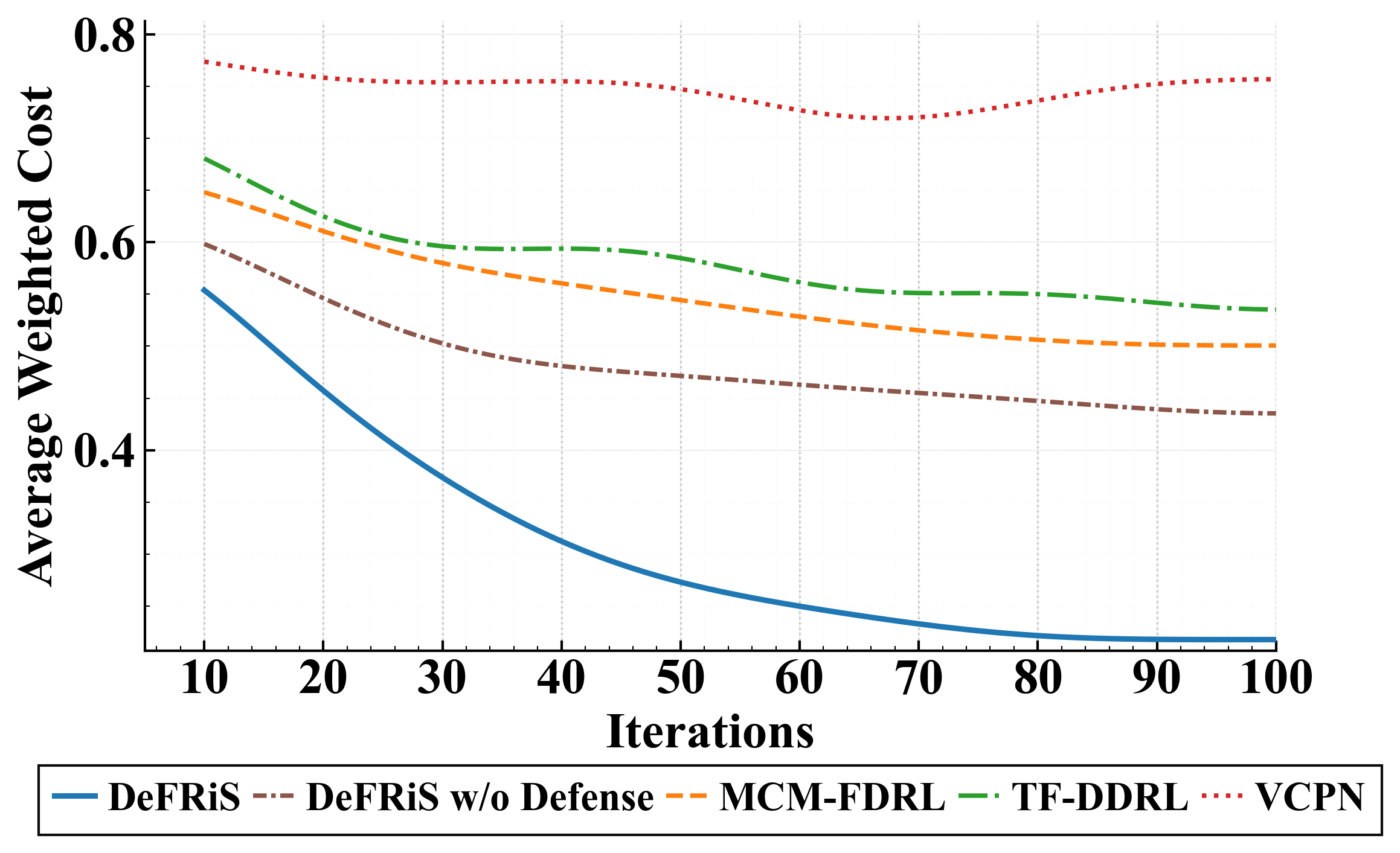}
    \caption{Robustness evaluation in adversarial environments.}
    \label{fig:robustness}
\end{figure}

To evaluate the robustness of DeFRiS in adversarial environments, we introduce 30\% malicious nodes into the 20-silo system, implementing three realistic threat scenarios: random noise injection (representing hardware failures) \cite{ahmadilivani2024systematic}, gradient reversal attacks (representing malicious behavior) \cite{huang2021evaluating}, and intermittent network disruptions (representing communication faults) \cite{yu2023communication}. We compare the performance of full DeFRiS, DeFRiS without anomaly detection (DeFRiS w/o Defense), and three baseline approaches involving cross-silo cooperation (MCM-FDRL, TF-DDRL, and VCPN) under attack. Since D3QN-SEIDEL trains each silo independently without cooperation mechanisms, it is immune to distributed attacks and thus excluded from this experiment. Figure~\ref{fig:robustness} presents the average weighted cost across 100 training iterations for all approaches.

DeFRiS demonstrates superior robustness under attack. Full DeFRiS converges from approximately 0.554 to approximately 0.218, with final performance degrading only 17.8\% compared to the benign scenario (0.185). In contrast, DeFRiS w/o Defense converges to approximately 0.435, suffering 135.1\% performance degradation compared to the benign scenario, validating the critical role of the anomaly detection mechanism. Baseline approaches exhibit severe vulnerability: MCM-FDRL achieves approximately 0.501, representing 162.3\% degradation compared to its benign performance (0.191); TF-DDRL reaches approximately 0.527, degrading 152.2\% from its benign performance (0.209); while VCPN barely improves during training (from around 0.774 to around 0.757, only 2.2\% reduction) with 74.8\% degradation from its benign performance (0.433), nearly losing all learning capability. This highlights that DeFRiS achieves over 8 times better performance stability compared to the best-performing baseline. These results demonstrate that DeFRiS's anomaly detection mechanism effectively identifies and filters corrupted parameters through gradient fingerprint-based similarity assessment and robust statistical thresholding (median and MAD), while baseline approaches lack robustness mechanisms, leading to training collapse or severe performance degradation in adversarial environments.

\section{Conclusions and Future Work}
\label{conclusions}
In this paper, we propose DeFRiS, a decentralized federated reinforcement learning framework designed to tackle the challenges of infrastructure heterogeneity, Non-IID workload shifts, and adversarial risks in silo-cooperative IoT scheduling. By integrating an action-space-agnostic policy for seamless cross-silo knowledge transfer, a GAE-enhanced local learning mechanism for stable convergence under sparse rewards, and a gradient fingerprint-based robust aggregation protocol, DeFRiS effectively synthesizes collective intelligence without a central coordinator. Extensive experiments on a real-world distributed testbed demonstrate that DeFRiS significantly outperforms state-of-the-art baselines, reducing average response time by 6.4\% and energy consumption by 7.2\% compared to the best-performing method while ensuring strict QoS guarantees. Notably, the framework exhibits exceptional resilience and scalability, achieving over 3 times better performance retention as the system scales and over 8 times better stability in adversarial environments.

Future work will focus on enhancing the communication efficiency and security of the framework. We plan to incorporate gradient compression and quantization techniques to alleviate bandwidth constraints in edge environments, while integrating advanced cryptographic mechanisms, such as Post-Quantum Cryptography (PQC) and Secure Multi-Party Computation (SMPC), to guarantee long-term cryptographic resilience and robust data confidentiality. Additionally, we intend to extend DeFRiS to support emerging data-intensive workloads, specifically investigating how silo-cooperative scheduling can optimize distributed inference and fine-tuning tasks for Large Language Models (LLMs) and next-generation generative AI applications.

\ifCLASSOPTIONcaptionsoff
  \newpage
\fi

\bibliographystyle{IEEEtran}
\bibliography{IEEEabrv,Bibliography}

\begin{thebibliography}{10}
\providecommand{\url}[1]{#1}
\csname url@samestyle\endcsname
\providecommand{\newblock}{\relax}
\providecommand{\bibinfo}[2]{#2}
\providecommand{\BIBentrySTDinterwordspacing}{\spaceskip=0pt\relax}
\providecommand{\BIBentryALTinterwordstretchfactor}{4}
\providecommand{\BIBentryALTinterwordspacing}{\spaceskip=\fontdimen2\font plus
\BIBentryALTinterwordstretchfactor\fontdimen3\font minus \fontdimen4\font\relax}
\providecommand{\BIBforeignlanguage}[2]{{%
\expandafter\ifx\csname l@#1\endcsname\relax
\typeout{** WARNING: IEEEtran.bst: No hyphenation pattern has been}%
\typeout{** loaded for the language `#1'. Using the pattern for}%
\typeout{** the default language instead.}%
\else
\language=\csname l@#1\endcsname
\fi
#2}}
\providecommand{\BIBdecl}{\relax}
\BIBdecl

\bibitem{chang2024digital}
X.~Chang, R.~Zhang, J.~Mao, and Y.~Fu, ``Digital twins in transportation infrastructure: an investigation of the key enabling technologies, applications, and challenges,'' \emph{IEEE Transactions on Intelligent Transportation Systems}, vol.~25, no.~7, pp. 6449--6471, 2024.

\bibitem{taleb2025survey}
I.~Taleb, J.-L. Guillaume, and B.~Duthil, ``A survey on services placement algorithms in integrated cloud-fog/edge computing,'' \emph{ACM Computing Surveys}, vol.~57, no.~11, pp. 1--36, 2025.

\bibitem{wu2024federated}
Z.~Wu, J.~Hou, Y.~Diao, and B.~He, ``Federated transformer: Multi-party vertical federated learning on practical fuzzily linked data,'' \emph{Advances in Neural Information Processing Systems}, vol.~37, pp. 45\,791--45\,818, 2024.

\bibitem{zhang2024high}
J.~Zhang, X.~Cheng, L.~Yang, J.~Hu, H.~Tian, and K.~Chen, ``High-performance hardware acceleration architecture for cross-silo federated learning,'' \emph{IEEE Transactions on Parallel and Distributed Systems}, vol.~35, no.~8, pp. 1506--1523, 2024.

\bibitem{li2025dynamic}
Y.~Li, J.-K. Hao, and L.~Song, ``Dynamic bin packing with heterogeneous dependent bins for regionless in geo-distributed clouds,'' \emph{IEEE Transactions on Computers}, 2025.

\bibitem{10887320}
S.~Tian, S.~Xiang, Z.~Zhou, H.~Dai, E.~Yu, and Q.~Deng, ``Task offloading and resource allocation based on reinforcement learning and load balancing in vehicular networking,'' \emph{IEEE Transactions on Consumer Electronics}, vol.~71, no.~1, pp. 2217--2230, 2025.

\bibitem{fan2024taking}
J.~Fan, K.~Wu, G.~Tang, Y.~Zhou, and S.~Huang, ``Taking advantage of the mistakes: Rethinking clustered federated learning for iot anomaly detection,'' \emph{IEEE Transactions on Parallel and Distributed Systems}, vol.~35, no.~6, pp. 862--876, 2024.

\bibitem{uddin2025systematic}
M.~P. Uddin, Y.~Xiang, M.~Hasan, J.~Bai, Y.~Zhao, and L.~Gao, ``A systematic literature review of robust federated learning: Issues, solutions, and future research directions,'' \emph{ACM Computing Surveys}, vol.~57, no.~10, pp. 1--62, 2025.

\bibitem{10623738}
X.~Li, W.~Huangfu, X.~Xu, J.~Huo, and K.~Long, ``Secure offloading with adversarial multi-agent reinforcement learning against intelligent eavesdroppers in uav-enabled mobile edge computing,'' \emph{IEEE Transactions on Mobile Computing}, vol.~23, no.~12, pp. 13\,914--13\,928, 2024.

\bibitem{yunis2024subwords}
D.~Yunis, J.~Jung, F.~Dai, and M.~Walter, ``Subwords as skills: Tokenization for sparse-reward reinforcement learning,'' \emph{Advances in Neural Information Processing Systems}, vol.~37, pp. 67\,663--67\,688, 2024.

\bibitem{li2023end}
H.~Li, C.~Dou, D.~Yue, G.~P. Hancke, Z.~Zeng, W.~Guo, and L.~Xu, ``End-edge-cloud collaboration-based false data injection attack detection in distribution networks,'' \emph{IEEE Transactions on Industrial Informatics}, vol.~20, no.~2, pp. 1786--1797, 2023.

\bibitem{10417755}
T.~Zeng, X.~Zhang, J.~Duan, C.~Yu, C.~Wu, and X.~Chen, ``An offline-transfer-online framework for cloud-edge collaborative distributed reinforcement learning,'' \emph{IEEE Transactions on Parallel and Distributed Systems}, vol.~35, no.~5, pp. 720--731, 2024.

\bibitem{qiu2020distributed}
X.~Qiu, W.~Zhang, W.~Chen, and Z.~Zheng, ``Distributed and collective deep reinforcement learning for computation offloading: A practical perspective,'' \emph{IEEE Transactions on Parallel and Distributed Systems}, vol.~32, no.~5, pp. 1085--1101, 2020.

\bibitem{jin2024collaborative}
D.~Jin, N.~Kannengiesser, S.~Rank, and A.~Sunyaev, ``Collaborative distributed machine learning,'' \emph{ACM Computing Surveys}, vol.~57, no.~4, pp. 1--36, 2024.

\bibitem{gecer2024federated}
M.~Gecer and B.~Garbinato, ``Federated learning for mobility applications,'' \emph{ACM Computing Surveys}, vol.~56, no.~5, pp. 1--28, 2024.

\bibitem{ye2023heterogeneous}
M.~Ye, X.~Fang, B.~Du, P.~C. Yuen, and D.~Tao, ``Heterogeneous federated learning: State-of-the-art and research challenges,'' \emph{ACM Computing Surveys}, vol.~56, no.~3, pp. 1--44, 2023.

\bibitem{wu2024topology}
J.~Wu, F.~Dong, H.~Leung, Z.~Zhu, J.~Zhou, and S.~Drew, ``Topology-aware federated learning in edge computing: A comprehensive survey,'' \emph{ACM Computing Surveys}, vol.~56, no.~10, pp. 1--41, 2024.

\bibitem{10380323}
X.~Chen, S.~Hu, C.~Yu, Z.~Chen, and G.~Min, ``Real-time offloading for dependent and parallel tasks in cloud-edge environments using deep reinforcement learning,'' \emph{IEEE Transactions on Parallel and Distributed Systems}, vol.~35, no.~3, pp. 391--404, 2024.

\bibitem{10989563}
J.~Tang, W.~Zhao, J.~Jin, Y.~Xiang, X.~Wang, and Z.~Zhou, ``Adaptive search and collaborative offloading under device-to-device joint edge computing network,'' \emph{IEEE Transactions on Mobile Computing}, vol.~24, no.~10, pp. 9852--9867, 2025.

\bibitem{10848209}
X.~Deng, H.~Yang, J.~Zhang, J.~Gui, S.~Lin, X.~Wang, and G.~Min, ``Task offloading in internet of vehicles: A drl-based approach with representation learning for dag scheduling,'' \emph{IEEE Transactions on Mobile Computing}, vol.~24, no.~6, pp. 5045--5060, 2025.

\bibitem{wang2024deep}
Z.~Wang, M.~Goudarzi, M.~Gong, and R.~Buyya, ``Deep reinforcement learning-based scheduling for optimizing system load and response time in edge and fog computing environments,'' \emph{Future Generation Computer Systems}, vol. 152, pp. 55--69, 2024.

\bibitem{11164488}
B.~Zhang, T.~Jing, Q.~Gao, X.~Li, and M.~Zhu, ``D3qn-enabled diversified-task co-offloading for synthetic-expense minimization in industrial internet of things (iiot),'' \emph{IEEE Transactions on Industrial Informatics}, vol.~21, no.~12, pp. 9491--9502, 2025.

\bibitem{10843979}
P.~Li, Z.~Xiao, H.~Gao, X.~Wang, and Y.~Wang, ``Reinforcement learning based edge-end collaboration for multi-task scheduling in 6g enabled intelligent autonomous transport systems,'' \emph{IEEE Transactions on Intelligent Transportation Systems}, vol.~26, no.~10, pp. 17\,624--17\,637, 2025.

\bibitem{10960753}
Q.~Chen, X.~Song, T.~Song, and Y.~Yang, ``Vehicular edge computing networks optimization via drl-based communication resource allocation and load balancing,'' \emph{IEEE Transactions on Mobile Computing}, vol.~24, no.~9, pp. 9222--9237, 2025.

\bibitem{11039641}
Y.~Liu, L.~Jiang, C.~Yuen, and Y.~Zhang, ``Vehicular computing power networks for iot-driven edge intelligence: Ma-ddpg-based robust task offloading and resource allocation,'' \emph{IEEE Internet of Things Journal}, vol.~12, no.~18, pp. 36\,868--36\,879, 2025.

\bibitem{10540320}
C.~Wu, Z.~Xu, X.~He, Q.~Lou, Y.~Xia, and S.~Huang, ``Proactive caching with distributed deep reinforcement learning in 6g cloud-edge collaboration computing,'' \emph{IEEE Transactions on Parallel and Distributed Systems}, vol.~35, no.~8, pp. 1387--1399, 2024.

\bibitem{zhang2024lsia3cs}
Z.~Zhang, F.~Zhang, Z.~Xiong, K.~Zhang, and D.~Chen, ``Lsia3cs: Deep reinforcement learning-based cloud-edge collaborative task scheduling in large-scale iiot,'' \emph{IEEE Internet of Things Journal}, vol.~11, no.~13, pp. 23\,917--23\,930, 2024.

\bibitem{wang2025tf}
Z.~Wang, M.~Goudarzi, and R.~Buyya, ``Tf-ddrl: A transformer-enhanced distributed drl technique for scheduling iot applications in edge and cloud computing environments,'' \emph{IEEE Transactions on Services Computing}, vol.~18, no.~2, pp. 1039--1053, 2025.

\bibitem{10618900}
M.~R. Raju, S.~K. Mothku, and M.~K. Somesula, ``Dmits: Dependency and mobility-aware intelligent task scheduling in socially-enabled vfc based on federated drl approach,'' \emph{IEEE Transactions on Intelligent Transportation Systems}, vol.~25, no.~11, pp. 17\,007--17\,022, 2024.

\bibitem{10949717}
X.~Chen, B.~Xiao, X.~Lin, Z.~Chen, and G.~Min, ``Multi-agent collaboration for vehicular task offloading using federated deep reinforcement learning,'' \emph{IEEE Transactions on Mobile Computing}, vol.~24, no.~9, pp. 8856--8871, 2025.

\bibitem{rockafellar2000optimization}
R.~T. Rockafellar, S.~Uryasev \emph{et~al.}, ``Optimization of conditional value-at-risk,'' \emph{Journal of Risk}, vol.~2, pp. 21--42, 2000.

\bibitem{sutton1984temporal}
R.~S. Sutton, \emph{Temporal credit assignment in reinforcement learning}.\hskip 1em plus 0.5em minus 0.4em\relax University of Massachusetts Amherst, 1984.

\bibitem{budennyy2022eco2ai}
S.~A. Budennyy, V.~D. Lazarev, N.~N. Zakharenko, A.~N. Korovin, O.~Plosskaya, D.~V. Dimitrov, V.~Akhripkin, I.~Pavlov, I.~V. Oseledets, I.~S. Barsola \emph{et~al.}, ``Eco2ai: carbon emissions tracking of machine learning models as the first step towards sustainable ai,'' in \emph{Doklady Mathematics}, vol. 106, no. Suppl 1.\hskip 1em plus 0.5em minus 0.4em\relax Springer, 2022, pp. S118--S128.

\bibitem{yang2022torchaudio}
Y.-Y. Yang, M.~Hira, Z.~Ni, A.~Astafurov, C.~Chen, C.~Puhrsch, D.~Pollack, D.~Genzel, D.~Greenberg, E.~Z. Yang \emph{et~al.}, ``Torchaudio: Building blocks for audio and speech processing,'' in \emph{Proceedings of the IEEE International Conference on Acoustics, Speech and Signal Processing}, 2022.

\bibitem{mcfee2015librosa}
B.~McFee, C.~Raffel, D.~Liang, D.~P. Ellis, M.~McVicar, E.~Battenberg, and O.~Nieto, ``librosa: Audio and music signal analysis in python.'' \emph{SciPy}, vol. 2015, pp. 18--24, 2015.

\bibitem{bradski2000opencv}
G.~Bradski, ``The opencv library,'' \emph{Dr. Dobb's Journal of Software Tools}, 2000.

\bibitem{king2009dlib}
D.~E. King, ``Dlib-ml: A machine learning toolkit,'' \url{http://dlib.net/}, pp. 1755--1758, 2009.

\bibitem{clark2015pillow}
A.~Clark and Contributors, ``Pillow (pil fork) documentation,'' \url{https://pillow.readthedocs.io/}, 2015.

\bibitem{loria2018textblob}
S.~Loria \emph{et~al.}, ``textblob documentation,'' \emph{Release 0.15}, vol.~2, no.~8, p. 269, 2018.

\bibitem{bird2006nltk}
S.~Bird, ``Nltk: the natural language toolkit,'' in \emph{Proceedings of the COLING/ACL Interactive Presentation Sessions}, 2006.

\bibitem{abadi2016tensorflow}
M.~Abadi, P.~Barham, J.~Chen, Z.~Chen, A.~Davis, J.~Dean, M.~Devin, S.~Ghemawat, G.~Irving, M.~Isard \emph{et~al.}, ``Tensorflow: a system for large-scale machine learning,'' in \emph{Proceedings of the USENIX Symposium on Operating Systems Design and Implementation (OSDI)}, 2016.

\bibitem{tflite2017}
{TensorFlow Team}, ``Tensorflow lite: On-device machine learning framework,'' \url{https://www.tensorflow.org/lite}, 2017.

\bibitem{easyocr2020}
{JaidedAI}, ``Easyocr: Ready-to-use ocr with 80+ supported languages,'' \url{https://github.com/JaidedAI/EasyOCR}, 2020.

\bibitem{gailly1995zlib}
J.-l. Gailly and M.~Adler, ``zlib: A massively spiffy yet delicately unobtrusive compression library,'' \url{https://www.zlib.net/}, 1995.

\bibitem{rfc1952}
P.~Deutsch, ``Gzip file format specification version 4.3,'' \url{https://tools.ietf.org/html/rfc1952}, RFC 1952, Internet Engineering Task Force, Tech. Rep., 1996.

\bibitem{lugaresi2019mediapipe}
C.~Lugaresi \emph{et~al.}, ``Mediapipe: A framework for perceiving and processing reality,'' in \emph{Proceedings of the Workshop on Computer Vision for AR/VR at IEEE Computer Vision and Pattern Recognition}, 2019.

\bibitem{wang2025reinfog}
Z.~Wang, M.~Goudarzi, and R.~Buyya, ``Reinfog: A deep reinforcement learning empowered framework for resource management in edge and cloud computing environments,'' \emph{Journal of Network and Computer Applications}, vol. 242, pp. 1--20, 2025.

\bibitem{ahmadilivani2024systematic}
M.~H. Ahmadilivani, M.~Taheri, J.~Raik, M.~Daneshtalab, and M.~Jenihhin, ``A systematic literature review on hardware reliability assessment methods for deep neural networks,'' \emph{ACM Computing Surveys}, vol.~56, no.~6, pp. 1--39, 2024.

\bibitem{huang2021evaluating}
Y.~Huang, S.~Gupta, Z.~Song, K.~Li, and S.~Arora, ``Evaluating gradient inversion attacks and defenses in federated learning,'' \emph{Advances in neural information processing systems}, vol.~34, pp. 7232--7241, 2021.

\bibitem{yu2023communication}
E.~Yu, D.~Dong, and X.~Liao, ``Communication optimization algorithms for distributed deep learning systems: A survey,'' \emph{IEEE Transactions on Parallel and Distributed Systems}, vol.~34, no.~12, pp. 3294--3308, 2023.

\end{thebibliography}

\vfill

\end{document}